\documentclass[10pt,journal]{IEEEtran}
\usepackage[cmex10]{amsmath}
\usepackage{amsthm}
\usepackage{amssymb}
\usepackage{makecell} 
\usepackage{mathtools}
\usepackage{enumitem}
\setlist[enumerate]{leftmargin=5mm, label=\roman*)}
\setlist[itemize]{leftmargin=0mm}
\usepackage{cite}
\usepackage{flafter}
\usepackage[table]{xcolor}
\usepackage{ifpdf}
\usepackage[pdftex]{graphicx}
\usepackage{lipsum}
\usepackage{url}
\usepackage{algorithm,algcompatible}
\usepackage{graphicx}
\usepackage[caption=true,font=small,labelfont=sf,textfont=sf]{subfig}
\usepackage{multirow}
\usepackage{array}
\usepackage{color}
\newcommand{\subparagraph}[1]{\vspace{0.5em}\noindent\textbf{#1.} }

\theoremstyle{definition}

\allowdisplaybreaks
\ifCLASSINFOpdf
\else
\fi
\begin{document}
\title{A Comprehensive Survey on Spectral Clustering with Graph Structure Learning}
\author{Kamal Berahmand, Farid Saberi-Movahed, Razieh Sheikhpour, Yuefeng Li, and Mahdi Jalili,~\IEEEmembership{Member,~IEEE}
\IEEEcompsocitemizethanks{\IEEEcompsocthanksitem Kamal Berahmand, Yuefeng Li are with School of Computer Sciences, Faculty of Science, Queensland University of Technology (QUT), Brisbane, Australia. (E-mail: kamal.berahmand@hdr.qut.edu.au, y2.li@qut.edu.au).
\IEEEcompsocthanksitem Farid Saberi-Movahed is with the Department of Applied Mathematics, Faculty of Sciences and Modern Technologies, Graduate University of Advanced Technology, Kerman, Iran (E-mail: f.saberimovahed@kgut.ac.ir).
\IEEEcompsocthanksitem Razieh Sheikhpour is with the Department of Computer Engineering, Faculty of Engineering, Ardakan University, P.O. Box 184, Ardakan, Iran (E-mail: rsheikhpour@ardakan.ac.ir).
\IEEEcompsocthanksitem Mahdi Jalili is with the School of Engineering, RMIT University, Melbourne, VIC 3000, Australia (E-mail: mahdi.jalili@rmit.edu.au).
\IEEEcompsocthanksitem Corresponding author: Razieh Sheikhpour.
}
%
}

\markboth{}%
{Shell \MakeLowercase{\textit{et al.}}: Bare Demo of IEEEtran.cls for Computer Society Journals}
\IEEEtitleabstractindextext{%
\begin{abstract}
Spectral clustering is a powerful technique for clustering high-dimensional data, utilizing graph-based representations to detect complex, non-linear structures and non-convex clusters. The construction of a similarity graph is essential for ensuring accurate and effective clustering, making graph structure learning (GSL) central for enhancing spectral clustering performance in response to the growing demand for scalable solutions. Despite advancements in GSL, there is a lack of comprehensive surveys specifically addressing its role within spectral clustering. To bridge this gap, this survey presents a comprehensive review of spectral clustering methods, emphasizing on the critical role of GSL. We explore various graph construction techniques, including pairwise, anchor, and hypergraph-based methods, in both fixed and adaptive settings. Additionally, we categorize spectral clustering approaches into single-view and multi-view frameworks, examining their applications within one-step and two-step clustering processes. We also discuss multi-view information fusion techniques and their impact on clustering data. By addressing current challenges and proposing future research directions, this survey provides valuable insights for advancing spectral clustering methodologies and highlights the pivotal role of GSL in tackling large-scale and high-dimensional data clustering tasks.
\end{abstract}

}
\maketitle
\IEEEdisplaynontitleabstractindextext
\IEEEpeerreviewmaketitle

\section{Introduction}\label{sec1}

 \IEEEPARstart{C}LUSTRING, is a fundamental technique in unsupervised learning, aimed at partitioning data points into distinct groups or clusters, such that points within a cluster are more similar to each other than to those in other clusters \cite{xue2024comprehensive, xie2023efficient, singh2024comprehensive}. Unlike supervised learning, clustering operates without predefined labels or categories, instead identifying inherent patterns and structures within the data. This makes it particularly useful for exploratory data analysis, where the objective is to uncover hidden patterns without prior assumptions about the data structure \cite{feng2023review}. Clustering is widely applied across various fields, including marketing \cite {benslama2020clustering}, social network analysis \cite {liu2019learning}, image segmentation \cite {liang2019image}, bioinformatics \cite {higham2007spectral}, anomaly detection \cite {li2021clustering}, and document categorization \cite {cai2005document}. It simplifies complex data, enhances understanding, and often serves as a preprocessing step for other machine learning tasks, such as classification.

\begin{figure}[!h]
\centering
\includegraphics[width=1.05\linewidth]{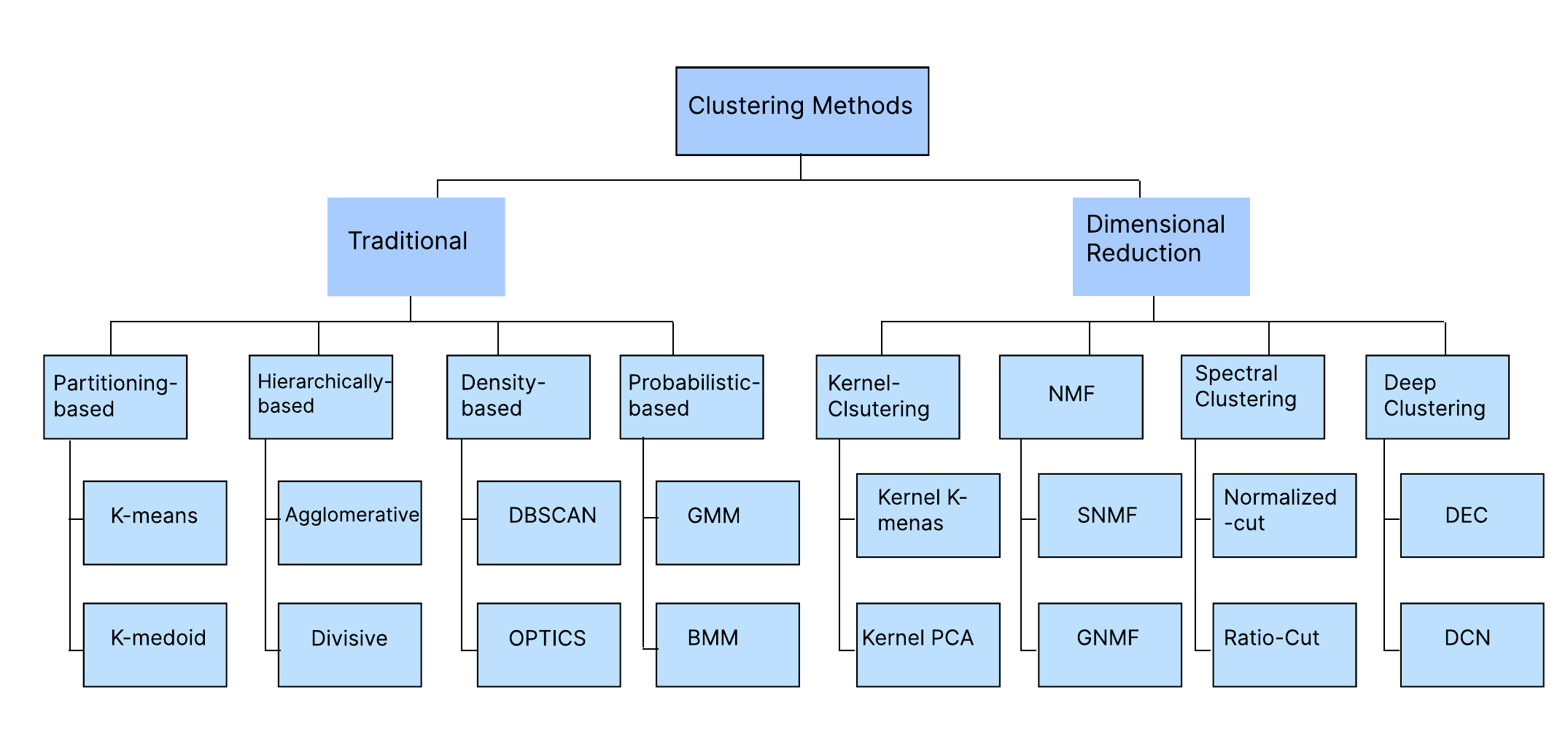}
\caption{Exploring Clustering Methods: From Traditional to Deep Learning Approaches.}
\label{figure1}
\end{figure}

Clustering methods can be broadly classified into traditional and dimensionality reduction-based approaches as illustrated in Fig. \ref{figure1}. Traditional methods include partitioning-based \cite{wang2017partition}, hierarchical \cite{shao2012synchronization}, density-based \cite{colomba2022density}, and probabilistic algorithms \cite{deng2018probabilistic}, each employing different strategies to group data. Partitioning-based methods, like K-means, divide data into a fixed number of clusters, each represented by a centroid \cite{wang2017partition}. Hierarchical methods, such as agglomerative and divisive clustering, build a cluster hierarchy by either merging smaller clusters (agglomerative) or splitting larger ones (divisive) \cite{murtagh2017algorithms, unger2020hierarchical}. Density-based methods, like DBSCAN, group data points based on regions of high density, identifying clusters of arbitrary shapes \cite{hahsler2019dbscan}. Probabilistic methods, such as Gaussian Mixture Models (GMM), use probabilistic models to represent data distributions and clusters \cite{deng2018probabilistic}.

While traditional methods are effective for lower-dimensional, well-structured datasets, they face limitations when applied to high-dimensional or complex data. In high-dimensional spaces, measuring distances between points becomes challenging, often resulting in poor clustering performance. Additionally, traditional methods often fail to capture non-convex shapes and intricate data structures. To address these limitations, dimensionality reduction-based clustering methods have emerged, reducing the number of features or dimensions and enabling clustering in a lower-dimensional space while preserving essential structural information.

Dimensionality reduction-based methods include Non-negative Matrix Factorization (NMF) \cite{berahmand2022graph}, spectral clustering \cite{von2007tutorial,liang2020multi}, kernel clustering \cite{filippone2008survey}, and deep clustering \cite{ren2024deep}. NMF is an effective dimensionality reduction technique used for clustering that factorizes the data matrix into two lower-dimensional non-negative matrices \cite{berahmand2022graph}. However, it may face challenges when dealing with more complex or non-linear data structures. Kernel clustering, including methods like Kernel K-means and Kernel PCA, handles non-linear relationships in the data by applying kernel functions \cite{filippone2008survey}. Spectral clustering utilizes graph theory, representing data points as nodes and their similarities as edges, while leveraging methods such as Ratio-cut \cite{hagen1992new} and Normalized-cut \cite{shi2000normalized}. Deep clustering integrates deep learning with clustering by using neural networks to learn lower-dimensional representations \cite{ren2024deep}. While deep clustering is powerful for large-scale, high-dimensional data, it requires significant computational resources and careful hyperparameter tuning.

Among dimensionality reduction techniques, spectral clustering stands out for its ability to handle complex data by identifying non-convex clusters and capturing non-linear structures through graph-based approaches. By representing data points as nodes in a graph and using graph-based embedding, spectral clustering partitions data based on connectivity and relationships. This flexibility makes spectral clustering applicable to a wide range of problems across various fields, particularly when combined with effective graph construction techniques. Spectral clustering is especially suitable for high-dimensional data, where spectral embedding reduces dimensionality while retaining essential structural information, mitigating challenges like the ``curse of dimensionality" and enabling reliable clustering of non-linear patterns. For large-scale datasets, anchor graph-based spectral clustering offers a scalable solution by using a subset of representative points, or anchors, to approximate relationships efficiently, preserving computational resources while maintaining clustering quality. Consequently, spectral clustering is both versatile and scalable, adaptable to high-dimensional and large-volume data applications, making it a powerful tool for complex clustering tasks
\cite{yang2017discrete,filippone2008survey}.

A critical factor in the success of spectral clustering is the construction of the similarity graph, which serves as the foundation for the entire process. This graph represents relationships between data points, with nodes corresponding to the points and edges representing their pairwise similarities. The quality of this graph significantly influences the spectral embedding and, consequently, the clustering results, as it directly affects how accurately the underlying structure of the data is captured \cite{yang2017discrete}. Common graph types used in spectral clustering include pairwise graphs \cite{wu2022effective}, anchor graphs \cite{qiang2021fast, yang2022efficient}, and hypergraphs \cite{zhang2020hypergraph, ahn2018hypergraph}, each offering distinct advantages depending on the nature of the data. These graphs can either be fixed, where the structure remains constant throughout the clustering process, or adaptive, where the graph structure is dynamically learned and updated during the process.

Despite the advancements made in spectral clustering, particularly in fields like image segmentation \cite{zhang2021image}, text classification \cite{janani2019text}, and industrial design \cite{langone2015ls}, a comprehensive survey on spectral clustering that focuses on graph structure learning (GSL) is still lacking. To address this gap, our survey presents an extensive review of spectral clustering, with particular emphasis on graph structure and its critical role in improving clustering accuracy. While a previous survey \cite{ding2024survey} provides an overview of spectral clustering, focusing on graph cuts, Laplacian matrices, and the clustering process, our survey delves into the more specific and crucial aspect of GSL. Thye previous survey emphasizes the mathematical foundations and applications of spectral clustering but do not extensively explore the importance of how graphs are constructed and how this impacts clustering performance. In contrast, our survey highlights the role of graph construction techniques, including pairwise, anchor, and hypergraph methods, in both fixed and adaptive forms. Additionally, we classify spectral clustering methods into single-view and multi-view approaches, examining their applications within both one-step and two-step frameworks. The distinction between these frameworks lies in whether clustering is treated as an independent step following the spectral embedding or is jointly optimized with it. We also provide a deeper examination of information fusion in multi-view spectral clustering, an area not covered in previous survey, offering new insights for improving clustering in high-dimensional, complex data.

Our contributions in this survey are outlined as follows:
\begin{itemize}
    \item For the first time, we present the most extensive and detailed survey on spectral clustering, with a particular emphasis on GSL, highlighting its importance for enhancing clustering accuracy.
    \item We provide a thorough review of various graph construction techniques, including pairwise, anchor, and hypergraphs, in both their fixed and adaptive forms. Additionally, we categorize spectral clustering methods into single-view and multi-view approaches, analyzing how different graph construction techniques are utilized in these approaches, and their application in both one-step and two-step clustering frameworks.
    \item We discuss information fusion techniques in multi-view spectral clustering, providing new insights into how integrating data from multiple sources can enhance clustering performance. This is particularly relevant for handling complex, heterogeneous, and high-dimensional data, making it an important contribution to advancing multi-view spectral clustering.
    \item We present an analysis of recent research advancements, emerging challenges, and potential future directions in spectral clustering and graph learning. This includes identifying promising areas for further research and the development of more advanced algorithms, contributing to the growth and evolution of the field.
\end{itemize}
The remainder of this paper is structured as follows: In Section \ref{SC}, we provide a detailed background on spectral clustering. Section \ref{SC_Stage} presents the main stages of spectral clustering, including graph structure learning, spectral embedding, and the partitioning phase. Section \ref{category} offers a comprehensive categorization of spectral clustering methods based on single-view and multi-view approaches, graph structure type, and one-step versus two-step frameworks, and explores different spectral clustering methods within each category. Section \ref{future} discusses future research directions, and finally, Section \ref{conclusion} presents the conclusion.

	\section{Background of Spectral Clustering} \label{SC}
	
	\subsection{Notations and Preliminaries}
	
	Table \ref{tab:notations} provides a summary of the essential notations and initial concepts utilized throughout this paper. The data matrix, defined as \( X = [x_1, x_2, \ldots, x_n] \), contains \( n \) samples, each denoted by \( x_i \in \mathbb{R}^{d \times 1} \), where \( i = 1, \ldots, n \). In the case of multi-view scenarios, the dataset is considered as \( \mathcal{D} = \{X^1, X^2, \ldots, X^V\} \), which includes \( V \) views. The data in the \( v \)-th view is represented by \( X^v \in \mathbb{R}^{d_v \times n} \), with \( n \) as the sample count and \( d_v \) as the number of features in the \( v \)-th view, for \( v = 1, \ldots, V \).
	
	\begin{table}[!h]
		\centering
		\caption{Summary of Key Notations}
		\label{tab:notations}
		\scalebox{0.8}{\begin{tabular}{|l|l|}
			\hline
			\textbf{Notation} & \textbf{Description} \\
			\hline
			$X = \{x_1, \ldots, x_n\}$ & Input data set with $n$ samples \\
			$x_i \in \mathbb{R}^{d\times 1}$ & The $i$-th sample, for $i=1,\ldots,n$ \\
			$\mathcal{G} = (\mathcal{V}, \mathcal{E})$ & Graph with the set of vertices $\mathcal{V}$ and the set of edges $\mathcal{E}$ \\
			$F$ & Cluster indicator matrix \\
			\( \mathcal{C} = \{\mathcal{C}_1, \ldots, \mathcal{C}_k\} \) & The set of clusters \\
$vol(\mathcal{C}_l)$ & Volume of a cluster $\mathcal{C}_l$, for $l=1,\ldots,k$\\
			$k$ & Number of clusters \\
			$L$ & Graph Laplacian matrix \\
			$L_{sym}$ & Symmetric graph Laplacian matrix \\
			$L_{rw}$ & Random Walk graph Laplacian matrix \\
			$L_h$ & Normalized hypergraph Laplacian matrix \\
			$W$ & Weighted adjacency matrix of $\mathcal{G}$ \\
			$D$ & Degree matrix of $\mathcal{G}$ \\
			$A$ & Similarity matrix \\
			$S$ & Similarity matrix (for adaptive methods) \\
			$P$ & Projection matrix \\
			$Z$ & Self-expressive matrix \\
			$I_p$ & Identity matrix of size $p$ \\
			$\|X\|_F$ & The Frobenius norm, $\|X\|_F = \sqrt{\sum_{i=1}^d \sum_{j=1}^n x_{ij}^2}$ \\
			$\|X\|_1$ & The $\ell_1$ norm, $\|X\|_1 = \sum_{i=1}^d \sum_{j=1}^n |x_{ij}|$ \\
			$\|X\|_*$ & The Nuclear norm of $X$ \\
			$\|X\|_{21}$ & The $\ell_{21}$ norm, $\|X\|_{21} = \sum_{i=1}^d \sqrt{\sum_{j=1}^n x_{ij}^2}$ \\
			$\|x_i\|_2$ & The Euclidean norm of $x_i$\\
			$X^T$ & The transpose operator of $X$ \\
			$Tr(\cdot)$ & The trace operator \\
			$\mathrm{Ind}$ & The set of indicator matrices \\
			\hline
		\end{tabular}}
	\end{table}

	\subsection{Graph-cut}
	Consider a graph \(\mathcal{G} = (\mathcal{V}, \mathcal{E}) \), where \( \mathcal{V} =\{v_1,v_2,\ldots,v_n\}\) represents the finite set of vertices and \( \mathcal{E} \subseteq \mathcal{V} \times \mathcal{V}\) denotes the set of edges, indicating the pairwise connections among vertices. Moreover, suppose that \( W \in \mathbb{R}^{n \times n} \) stands for the weighted adjacency matrix of $\mathcal{G} $ and \( D \) for its degree matrix. The goal of a graph-cut algorithm is to identify a partition \( \mathcal{C} = \{\mathcal{C}_1, \mathcal{C}_2, \ldots, \mathcal{C}_k\} \) of the vertex set \( \mathcal{V} \) such that $\bigcup_{i=1}^k \mathcal{C}_i = \mathcal{V}$, ensuring that the union of all clusters encompasses the entire vertex set \cite{von2007tutorial}. Furthermore, $\mathcal{C}_i \cap \mathcal{C}_j = \emptyset$ for all $i \neq j$, which guarantees that the clusters are disjoint, so each vertex belongs to only one cluster. Two effective formulations of the graph-cut problem are Ratio-cut (R-cut) \cite{hagen1992new} and Normalized-cut (N-cut) \cite{shi2000normalized}, which differ in how they handle cluster balance. For further explanation:
	
	\begin{enumerate}[leftmargin=5mm, label=(\roman*)] 
		
		\item \textbf{R-cut:}
		The R-cut approach divides the graph into the \( k \) clusters \( \mathcal{C}_1, \mathcal{C}_2, \ldots, \mathcal{C}_k \) by minimizing the cut value for each cluster, normalized by the number of nodes within it. This can be expressed as \cite{hagen1992new}:
		\begin{equation}
			\min_{\mathcal{C}_1, \ldots, \mathcal{C}_k}Rcut(\mathcal{C}_1, \ldots, \mathcal{C}_k) = \sum_{l=1}^k \frac{cut(\mathcal{C}_l, \bar{\mathcal{C}_l})}{|\mathcal{C}_l|},
		\end{equation}
		where \( cut(\mathcal{C}_l, \bar{\mathcal{C}_l}) \) indicates the total edge weight connecting the nodes in the cluster \(\mathcal{C}_l \) to those outside it, and \( |\mathcal{C}_l| \) is the number of nodes in \( \mathcal{C}_l \).
		
		Let us also assume that \( L = D - W \) represents the graph Laplacian matrix, and that \(f_l \) denotes the cluster indicator vectors with $n$ elements, defined as \(f_{il} = \frac{1}{\sqrt{|\mathcal{C}_l|}} \) if \( v_i \in \mathcal{C}_l \); otherwise, \(f_{il} = 0 \), for \( i = 1, \ldots, n \) and \( l = 1, \ldots, k \). Accordingly, the objective function of the R-cut problem can be restated as:
		\begin{equation}
			Rcut(\mathcal{C}_1, \ldots, \mathcal{C}_k) = \sum_{l=1}^k \frac{f_l^T L f_l}{f_l^T f_l}.
		\end{equation}
		With this description, the R-cut problem is formulated as:
		\begin{equation}
			\min_{\mathcal{C}_1, \ldots, \mathcal{C}_k} Tr(F^TLF),
		\end{equation}
		where \(F= [f_1,\ldots,f_k]\) is the cluster indicator matrix and $F^TF=I_k$.
		
		\item \textbf{N-cut:}
		In contrast, the N-cut method adjusts each cut value according to the overall connectivity of the cluster, defined as the cluster volume, represented by \( vol(\mathcal{C}_l) \), where \( vol(\mathcal{C}_l) = \sum_{v_i \in \mathcal{C}_l} d_i \), and \( d_i \) is the degree of the \( i \)-th node.
		The N-cut problem is formulated as \cite{shi2000normalized}:
		\begin{equation}\label{dnsc0}
			\min_{\mathcal{C}_1, \ldots, \mathcal{C}_k}Ncut(\mathcal{C}_1, \ldots, \mathcal{C}_k) = \sum_{l=1}^k \frac{cut(\mathcal{C}_l, \bar{\mathcal{C}_l})}{vol(\mathcal{C}_l)}.
		\end{equation}
		By minimizing \( Ncut \), this approach promotes clusters that are densely connected internally while being loosely connected to other clusters. Similar to R-cut, the N-cut problem is also formulated using the Laplacian matrix as:
		\begin{equation}
			\min_{\mathcal{C}_1, \ldots, \mathcal{C}_k} \text{Tr}(F^T L F),
		\end{equation}
		where \(F= [f_1, \ldots,f_k] \) represents the cluster indicator matrix. The components of \(f_l \) are defined such that \(f_{il} = \frac{1}{\sqrt{vol(\mathcal{C}_l)}} \) if \( v_i \in \mathcal{C}_l \); otherwise, \(f_{il} = 0 \) for \( i = 1, \ldots, n \) and \( l = 1, \ldots, k \). Furthermore, it follows that \(F^T D F = I_k \).
	\end{enumerate}
	
	Although both the R-cut and N-cut are fundamental in graph clustering, their inherent NP-hardness arises from the discrete nature of the cluster indicator matrix \( F \) \cite{von2007tutorial}. To address this challenge, a relaxation technique can be employed that allows \( F \) to assume real values, transforming the discrete optimization objectives into continuous forms that can be efficiently solved using the eigenvalue decomposition. Specifically, the continuous relaxations
	can be presented through the following formulations:
	
	\textbf{$\bullet$ R-cut relaxation:}
	\begin{equation}
		\min_{F\in \mathbb{R}^{n \times k},\, F^T F = I_k} Tr(F^T LF).
	\end{equation}
	
	\textbf{$\bullet$ N-cut relaxation:}
	\begin{equation}
		\min_{F\in \mathbb{R}^{n \times k},\,F^T DF = I_k} Tr(F^T LF).
	\end{equation}
	
	\section{Main Stages of Spectral Clustering} \label{SC_Stage}
\begin{figure}[!h]
		\centering
		\includegraphics[width=0.48\textwidth]{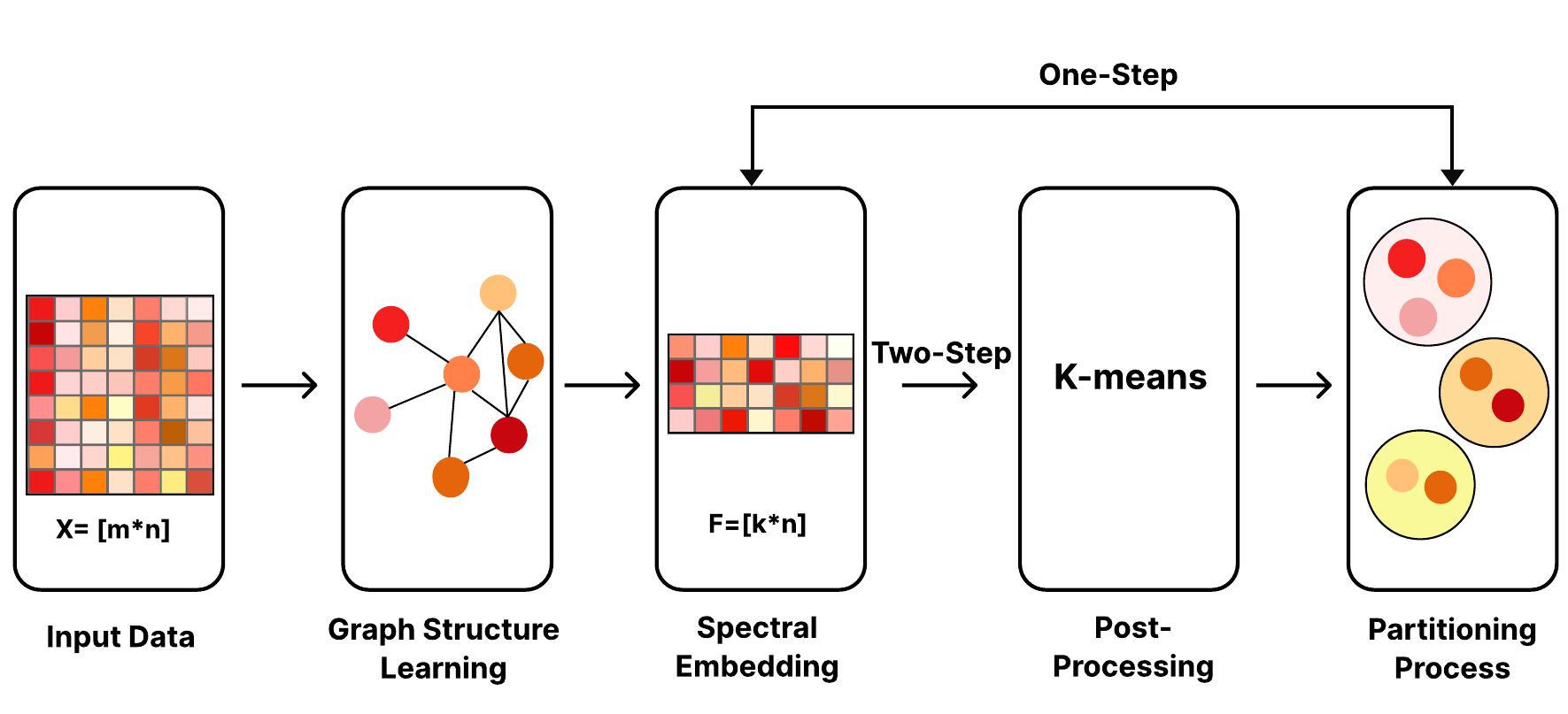}
		\caption{Systematic Overview of Spectral Clustering Framework.}
		\label{figure 2}
	\end{figure}

	Spectral clustering has emerged as a sophisticated graph-cut methodology in machine learning, distinguished by its ability to capture intrinsic manifold structures through the eigenspectrum analysis of graph Laplacian matrices \cite{xue2024comprehensive}. The mathematical foundation of spectral clustering integrates graph theory and manifold learning principles, enabling effective handling of complex, nonlinear data distributions while maintaining theoretical guarantees and computational efficiency in the clustering process. As illustrated in Fig. \ref{figure 2}, spectral clustering transcends traditional clustering methods by establishing the clustering objective into three theoretically-grounded stages: graph structure learning for optimal relationship modeling \cite{ding2024survey}, spectral embedding through eigendecomposition for manifold structure revelation \cite{pang2018spectral}, and partitioning process for final cluster assignment \cite{huang2013spectral}. Each of these stages will be discussed in detail in the following sections.
	
	\subsection{Graph Structure Learning}
	Graph structure learning (GSL) constitutes a fundamental paradigm in machine learning, focusing on the optimization of graph representations from empirical data \cite{kang2021structured}. As an essential component across multiple domains including Graph Neural Networks (GNNs) \cite{zhu2021deep}, manifold learning \cite{li2019survey}, and graph-based semi-supervised learning \cite{chong2020graph}, GSL stands as a cornerstone methodology for representing complex data relationships. Within the spectral clustering framework, GSL serves as a crucial component, where the quality of the learned graph structure directly determines the efficacy of subsequent spectral embedding and partitioning phase.
	
\begin{figure}[!h]
		\centering
		\includegraphics[width=0.5\textwidth]{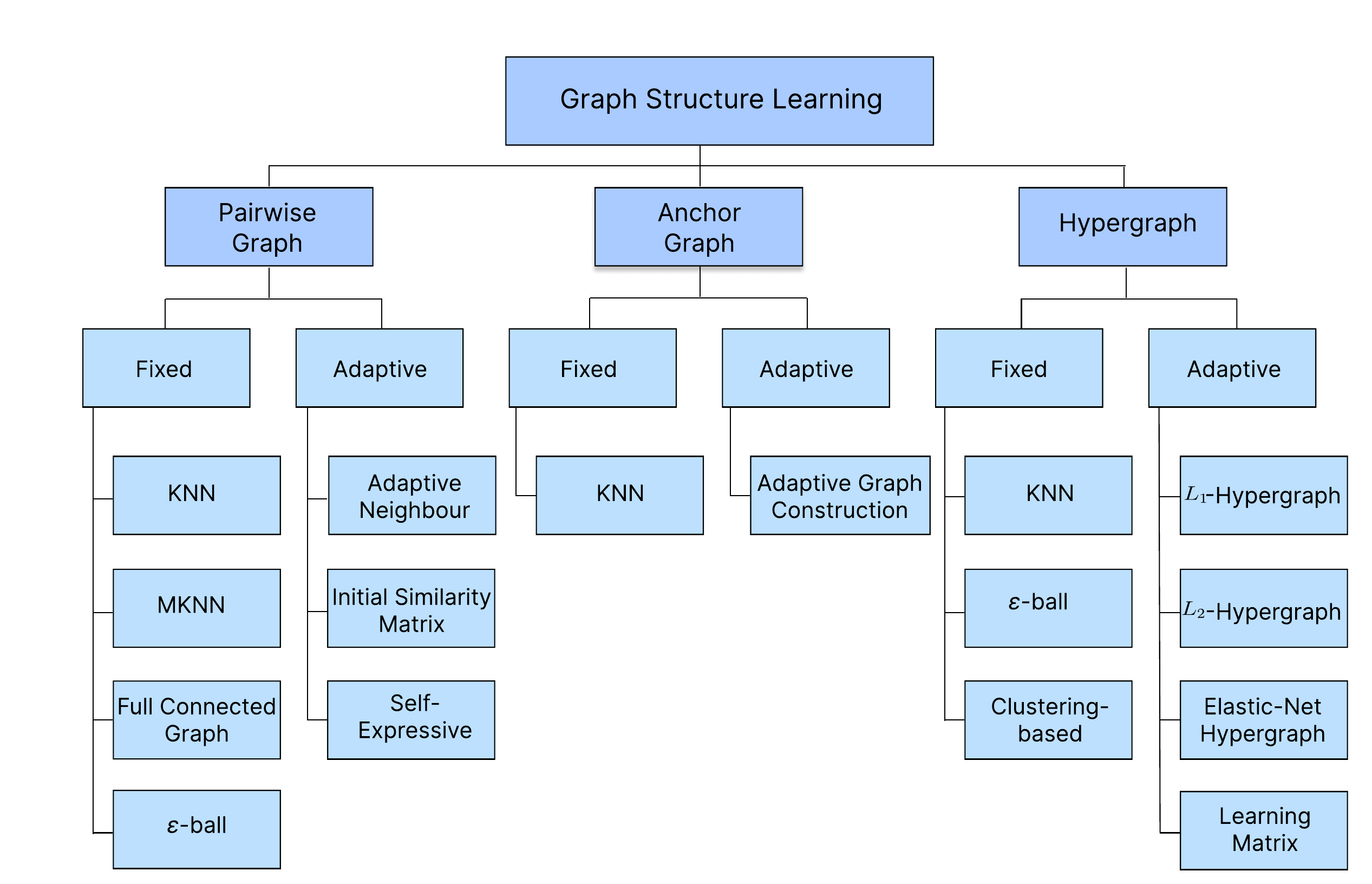}
		\caption{Graph Structure Learning Methods.}
		\label{figure 25}
	\end{figure}

	The taxonomy of GSL methodologies in spectral clustering encompasses two paradigms, fixed and adaptive, each offering distinct approaches to graph construction. Fixed methodologies employ predetermined protocols and parameters for graph construction, which enhances computational efficiency and generalization across datasets. In contrast, adaptive methodologies dynamically optimize the graph structure by designing a tailored optimization problem, enabling them to capture dataset-specific characteristics and provide greater representational flexibility for complex topological relationships. Selecting an appropriate methodology has a significant impact on clustering performance, as the resulting graph structure supports spectral embedding and subsequent partition optimization. As illustrated in Fig. \ref{figure 25}, GSL can be systematically categorized into three principal architectures: pairwise graphs, anchor-based graphs, and hypergraphs, each of which includes both fixed and adaptive methodological variants.
	
	\subsubsection{Pairwise Graph}
	The foremost and undeniably crucial step in spectral clustering involves constructing a similarity matrix. This matrix must reflect the pairwise similarities among the data points, meaning that points within the same cluster should exhibit strong connections with high similarity values, unlike points belonging to different clusters. When constructing pairwise graphs from a dataset with the pairwise similarities, the primary objective is to capture either the local neighborhood relationships between data points or the broader, global structure of the dataset. The methodologies for constructing pairwise graphs can be systematically categorized into two approaches: fixed \cite{ng2001spectral} and adaptive \cite{nie2014clustering}, each of which is explained below.
	
	\paragraph {Fixed Methods}
	Suppose that the set of data points to be partitioned into \( k \) subsets is indicated by \(\{x_1, \ldots, x_n\} \). To facilitate spectral clustering, the data must first be represented in the form of a similarity graph \( \mathcal{G} = (\mathcal{V}, \mathcal{E}) \). In this graph, each vertex \( v_i \) is associated with a corresponding data point \( x_i \), for \( i = 1, \ldots, n \), while \( \mathcal{E} \) represents the edges that connect the vertices. Furthermore, an \( n \times n \) similarity matrix \( A = [a_{ij}] \) is established, in which the values of \( a_{ij} \) represent the pairwise similarities between the data points \( x_i \) and \( x_j \).
	In pairwise graph construction, fixed methodologies refer to the approaches in which the graph structure is determined by fixed criteria instead of being dynamically adjusted to the data \cite{ng2001spectral}. These approaches provide computational efficiency while maintaining consistent structural properties across different datasets. The construction of a fixed-based pairwise graph is systematically performed through four primary principles \cite{von2007tutorial}:

	\begin{enumerate}[leftmargin=5mm, label=(\roman*)] 
		
		\item \textbf{K-Nearest Neighbors (KNN):} It links each data point to its \(K\) nearest neighbors based on a specified distance metric,such as the Euclidean distance. Consequently, a graph is created where the edges denote the closest \(K\) points to each node.
		
		\item \textbf{Mutual KNN (MKNN):} It builds upon KNN by creating an edge between two points only if each point is included in the KNN of the other. This guarantees a reciprocal connection, thus strengthening the relationships represented in the graph.
		
		\item \textbf{Fully Connected:} It describes a complete graph in which every distinct pair of nodes is interconnected by an edge, leading to a dense network of links.
		
		\item \textbf{$\varepsilon$-Ball:} This refers to a spherical area surrounding a point in a metric space, including the Euclidean metric space, characterized by a radius \( \varepsilon \). It encompasses all points that are within a distance of \( \varepsilon \) from the specified point, forming edges between the central point and all points contained in the $\varepsilon$-ball.
		
	\end{enumerate}
	
	Each of the methods described above constructs the similarity matrix \( \mathbf{A} = [a_{ij}] \) based on different criteria, including \cite{von2007tutorial}:
	
	\begin{enumerate}[leftmargin=5mm, label=(\roman*)]
		\item \textbf{Binary Similarity:} In this approach, the pairwise similarity \( a_{ij} \) is defined as \( a_{ij} = 1 \) if the data points \( {x}_i \) and \({x}_j \) are linked to each other, while \( a_{ij} = 0 \) otherwise.
		
		\item \textbf{Cosine Similarity:} This criterion evaluates the cosine of the angle formed by two data points. It is computed as
		$a_{ij} = \frac{{x}_i^T{x}_j}{\|{x}_i\|_2 \|{x}_j\|_2}$.
		
		\item \textbf{Gaussian Kernel:} This technique establishes pairwise similarity through the Gaussian kernel function, calculated as
		$ a_{ij} = \exp\left(-\frac{\|{x}_i - {x}_j\|_2^2}{2\sigma^2}\right)$, where \( \sigma \) is the bandwidth parameter that regulates the width of the Gaussian function.
	\end{enumerate}

	\paragraph{Adaptive Methods}
	
	Adaptive methods dynamically learn the graph structure based on data characteristics during the clustering process \cite{zhu2017one}. Unlike fixed methods, these approaches optimize the graph structure to better reflect intrinsic data relationships and adapt to local data distributions. This flexibility enables more accurate representation of complex data structures. The main adaptive approaches include:

	\subparagraph{Adaptive Neighbor}
	Compared to KNN-based graph construction, the adaptive neighbor approach provides a significant advancement by simultaneously learning both the neighbors of each data point and their respective weights from the original data. This adaptive mechanism selects optimal neighbors to ensure that points in closer proximity have stronger relationships. A particularly promising model in this context is the Clustering with Adaptive Neighbors (CAN) method \cite{nie2014clustering}, which formalizes this adaptive process within a probabilistic optimization framework, introduced as:
	\begin{align}
		&\min_{s_{ij}} \sum_{i,j=1}^n (\Vert x_i - x_j\Vert_2^2 s_{ij} + \gamma s_{ij}^2)\label{can1}\\
		&\text{s.t.}\quad\sum_{j=1}^n s_{ij} =1, 0 \leq s_{ij} \leq 1,\nonumber
	\end{align}
	where \( \gamma \) is the regularization parameter, and the matrix \( S = [s_{ij}] \in \mathbb{R}^{n \times n} \) represents neighbor assignments, with each \( s_{ij} \) indicating the probability that the data sample \( x_j \) is a neighbor of \( x_i \). 	The effectiveness of the CAN framework is further improved by incorporating a projection matrix \( P\in\mathbb{R}^{d\times m}\), where $m\leq d$, to tackle the challenges associated with clustering high-dimensional data. This novel approach, referred to as Projected CAN (PCAN) \cite{nie2014clustering}, seeks to optimize both dimensionality reduction and neighborhood learning simultaneously through the following problem:
	{\begin{align}
			&\min_{s_{ij},P} \sum_{i,j=1}^n (\Vert P^T x_i - P^T x_j\Vert_2^2 s_{ij} + \gamma s_{ij}^2)\label{pcan1}\\
			&\text{s.t.}\quad P\in\mathbb{R}^{d\times m},\, \sum_{j=1}^n s_{ij} =1, 0 \leq s_{ij} \leq 1,\, P^T B P = I_m,\nonumber
		\end{align}
		where \( B \) represents the total scatter matrix. Additionally, the imposition of the orthogonality constraint \( P^T B P = I_m \) guarantees statistical uncorrelatedness of the data in the subspace.
		
		\subparagraph{Initial Similarity Matrix}
		This approach iteratively refines an initial similarity matrix \( A \in \mathbb{R}^{n \times n} \) to achieve more accurate similarity relationships \cite{nie2016constrained}. It specifically employs two distinct distance constraints--the Frobenius norm and the \( L_1 \) norm--to enhance the initial similarity matrix \( A \) into a refined similarity matrix \( S \in \mathbb{R}^{n \times n} \). Consequently, this learning framework is expressed as follows:
		
		\textbf{$\bullet$ The Frobenius norm-based problem:}
		\begin{align}
			&\min_{S} \Vert S - A\Vert_F^2\label{ism1}\\
			&\text{s.t.}\quad S = [s_{ij}]\in\mathbb{R}^{n\times n}, \sum_{j=1}^n s_{ij} =1, 0 \leq s_{ij} \leq 1.\nonumber
		\end{align}

		\textbf{$\bullet$ The $L_1$ norm-based problem:}
		\begin{align}
			&\min_{S} \Vert S - A\Vert_1\label{ism2}\\
			&\text{s.t.}\quad S = [s_{ij}]\in\mathbb{R}^{n\times n}, \sum_{j=1}^n s_{ij} =1, 0 \leq s_{ij} \leq 1.\nonumber
		\end{align}
		Here, the constraints $\sum_{j=1}^n s_{ij} =1$ and $0 \leq s_{ij} \leq 1$ guarantee proper probability distributions. Moreover, the choice of different norm minimizations introduces versatility in adjusting the refinement approach: the Frobenius norm focuses on reducing the Euclidean distance between \( S \) and \( A \), leading to smoother updates in similarity, whereas the \( L_1 \) norm promotes sparsity, which can produce a clearer and more sparse similarity structure in \( S \).

		\subparagraph{Self-expressive}
		Self-expressive learning is an advanced approach for identifying intrinsic relationships between data points, especially useful when the data is situated across several subspaces \cite{elhamifar2013sparse}. This technique creates a self-expressive matrix \( Z \), which illustrates how each data point can be optimally reconstructed using the other points in the dataset, thereby uncovering the underlying subspace structures through a global representation. The self-expressive framework is fundamentally expressed as:
		\begin{align}
			&\min_{Z} \Vert X - f(X)Z\Vert_l + \Omega(X, Z)\label{selfex1}\\
			&\text{s.t.}\quad\quad Z \in \mathbb{R}^{n \times n},\, Z \in \mathcal{C},\nonumber
		\end{align}
		where $\Vert\cdot\Vert_l$ denotes a specified matrix norm, \( f(X) \) represents a matrix function (typically defined as \( f(X) = X \)), \( \Omega(X,Z) \) is a regularization term, and \( \mathcal{C} \) defines the constraint set. A key outcome of Problem \eqref{selfex1} is the formulation of a similarity matrix \( S \), expressed as $S = (|Z| + |Z^T|)/2$.
		This formulation aids in uncovering intrinsic subspace structures via the reconstruction term and different regularization methods. Furthermore, the symmetrization of \( Z \) ensures a well-defined similarity measure, which can be valuable for effective clustering applications.

		\subsubsection{Anchor Graph}
		Anchor graphs present a way to construct a graph by selecting a smaller, representative subset of nodes, known as ``anchors," rather than incorporating every node from a large dataset \cite{nie2023fast}. These anchors facilitate efficient information propagation across the graph. Centering on these anchors enables the graph to keep core relationships and structures intact without processing all nodes, improving computational speed and making analysis more manageable. The construction of an anchor graph framework involves three main stages: anchor point selection, anchor graph construction, and similarity matrix generation.
		
		\paragraph{Anchor Point Selection}
		Various strategies can be employed to select a set of $m$ anchor points from a dataset containing $n$ samples, where $m \ll n$. Some of the most common examples include \cite{nie2020unsupervised,nie2023fast}:
		
		\begin{enumerate}[leftmargin=5mm, label=(\roman*)]
			\item \textbf{Random Selection:} It selects a set of $m$ anchor points arbitrarily, offering simplicity and low computational cost, but may produce unrepresentative anchors, leading to weaker clustering results.
			
			\item \textbf{K-means (KM):} It selects anchor points by clustering the data into \( m \) groups and using the centroids as anchors. This method improves anchor representativeness by focusing on denser data regions but requires multiple iterations and is sensitive to initial cluster centers.
			
			\item \textbf{K-means++ (KM++):} It refines KM initialization by randomly selecting the first anchor and choosing subsequent points based on their squared distance from the nearest selected anchor, enhancing diversity and improving clustering performance.

			\item \textbf{Balanced K-Means via Hierarchical K-Means (BKHK):}
			The BKHK algorithm creates representative anchors for large datasets by employing a balanced binary tree structure. It repeatedly divides the data into two equal clusters using a binary k-means approach, which maintains equal cluster sizes. The algorithm minimizes a distance-based formulation related to cluster centers, assigning samples to clusters according to the smallest distance differences. This hierarchical balanced method is a key feature of the BKHK algorithm.
		\end{enumerate}

		\paragraph{Anchor Graph Construction}
		Anchor-based graph construction is based on two principal approaches \cite{wang2023efficient}: fixed methods with preset rules and adaptive techniques that dynamically determine connection strengths.
		
		\subparagraph{Fixed Construction}
		The fixed construction of an anchor graph is a structured methodology that involves two sequential steps: neighborhood identification and edge weight assignment. Given data points $\{x_1, \ldots, x_n\}$ and anchor points $\{u_1, \ldots, u_m\}$, the process is detailed as follows \cite{wang2023efficient}:
		
		\begin{enumerate}[leftmargin=5mm, label=(\roman*)]
			\item \textbf{ Neighborhood Identification:}
			Each data point $x_i$ is connected to its $r$ nearest anchor points through KNN, establishing the fundamental topology of the anchor graph.
			
			\item \textbf{Edge Weight Assignment:}
			The weight measure $w_{ij}$ between each data point $x_i$ and anchor point $u_j$, where $i = 1, \ldots, n$ and $j = 1, \ldots, m$, is calculated based on one of the following three main schemes \cite{wang2023efficient}:
			
			\begin{itemize}
				\item \textbf{Binary Weight:}
				Assigns $w_{ij} = 1$ if $u_j$ is among the $r$ nearest anchors of $x_i$, and $w_{ij} = 0$ otherwise.
				
				\item \textbf{Polynomial Kernel:}
				Defines $w_{ij}$ as $w_{ij} = ((x_i)^T u_j)^d$ when $u_j$ is one of the $r$ nearest anchors to $x_i$, and $w_{ij} = 0$ otherwise. Here, the variable $d$ indicates the degree of the polynomial used in the kernel function.
				
				\item \textbf{Gaussian Kernel:}
				Computes $w_{ij}$ as $w_{ij} = \mathrm{exp}(-\frac{\|x_i - u_j\|_2^2}{2\sigma^2})$ if $u_j$ is among the nearest $r$ anchors of $x_i$, setting $w_{ij} = 0$ otherwise.
			\end{itemize}
		\end{enumerate}

		\subparagraph{Adaptive Construction}
		In anchor graph construction, adaptive approaches extend the principles of adaptive neighbor methods used in pairwise graphs. The main objective is to learn the neighbors and weights of each data sample based on its relationships with anchor points, aiming to enhance the representation of both local and global data structures. The optimization problem for adaptive anchor graph learning on each data sample $x_i$ (for $i=1,\ldots,n$) is defined as \cite{wang2023efficient}:
		\begin{align}
			&\min_{w_{ij}} \sum_{j=1}^m (\Vert x_i - u_j\Vert_2^2 w_{ij} + \gamma w_{ij}^2)\label{AGC1}\\
			&\text{s.t.}\quad\sum_{j=1}^n w_{ij} =1,\, w_{ij}>0,\nonumber
		\end{align}
		where $w_{ij}$ encodes the learned weight between the data sample $x_i$ and the anchor point $u_j$, and $\gamma$ is a regularization parameter.

		\paragraph{Similarity Matrix Generation}
		The generation of similarity matrices from anchor-based representations is a crucial step in establishing effective relationships between the data and anchor points.
		Two primary approaches have emerged for computing the similarity matrix within an anchor graph framework: the Gram matrix formulation \cite{cai2014large} and the bipartite graph construction \cite{nie2023fast}. Each of them demonstrates theoretical elegance in representing complex data relationships while maintaining computational efficiency. For additional details,
		
		\begin{enumerate}[leftmargin=5mm, label=(\roman*)]
			\item \textbf{Gram Matrix Formulation:} The inner product is a key concept in linear algebra for assessing the similarity between two entities. When two entities, such as vectors, have a high inner product, it means they point in a similar direction or share similar features. A practical application of inner products in representation learning involves constructing Gram matrices, which contain the inner products between all pairs of data points. This Gram matrix can be regarded as a similarity matrix, indicating how each data point relates to others. Let \(W = [w_{ij}] \in \mathbb{R}^{n \times m}\) be a weight matrix constructed between the data and anchors points, where each \(w_{ij}\) indicates the similarity between the \(i\)-th data sample and the \(j\)-th anchor point, for \(i = 1, \ldots, n\) and \(j = 1, \ldots, m\). Then the associated Gram matrix $S\in \mathbb{R}^{n \times n}$, encoding the pairwise similarities induced by the anchor-based representation, is established as \cite{cai2014large}:
			\begin{equation}
				S = WW^T.
			\end{equation}
			
			\item \textbf{Bipartite Graph Construction:}
			Constructing a bipartite graph serves as an effective method for modeling the relationships between samples and anchors \cite{zhu2021unsupervised}. To illustrate this, let \(\mathcal{G}\) be a bipartite graph consisting of \(n + m\) nodes, with the left-side nodes corresponding to the \(n\) data samples and the right-side nodes representing the \(m\) anchor points. Additionally, assume that \(W = [w_{ij}] \in \mathbb{R}^{n \times m}\) denotes the weight matrix for \(\mathcal{G}\). Based on this framework, the associated similarity matrix \(S \in \mathbb{R}^{(n + m) \times (n + m)}\) is defined as \cite{yang2023fast}:
			\begin{equation}
				S = \begin{bmatrix} 0 & W \\ W^T & 0 \end{bmatrix}.
			\end{equation}
		\end{enumerate}

		\subsubsection{Hypergraph}
		Assume that \(\mathcal{G} = (\mathcal{V}, \mathcal{E}, w) \) presents a weighted hypergraph, where \(\mathcal{V}\) is the vertex set, \(\mathcal{E}\) is the set of hyperedges, and \( w \) is the weight vector for the hyperedges, where each element represents the positive weight assigned to each hyperedge \cite{zhou2006learning}. The degree of each vertex \( v \in \mathcal{V} \) and the degree of each hyperedge \( e \in \mathcal{E} \) are respectively defined as $d(v) = \sum_{e \in \mathcal{E}} w(e) \, h(v, e)$ and $d(e) = \sum_{v \in \mathcal{V}} h(v, e)$, where \( w(e) \) represents the weight assigned to each hyperedge \( e \), and \( h(v, e) \) is an element of the incidence matrix \( H \in \mathbb{R}^{|\mathcal{V}| \times |\mathcal{E}|} \), which describes the connections between vertices and hyperedges. The entries of \( H \) are specified by \( h(v, e) = 1 \) if $v \in e$; otherwise, \( h(v, e) = 0 \).
		Let \( W \in \mathbb{R}^{|\mathcal{E}| \times |\mathcal{E}|} \) be the diagonal matrix formed from the weight vector \( w \) for the hyperedges of \( G \). Let us also define \( D_v \in \mathbb{R}^{|\mathcal{V}| \times |\mathcal{V}|}\) and \( D_e \in \mathbb{R}^{|\mathcal{E}| \times |\mathcal{E}|}\) as the diagonal matrices constructed from the vertex degrees \( d(v) \) and the hyperedge degrees \( d(e) \), respectively. The normalized Laplacian matrix for the hypergraph is expressed as
		$L = I - D_v^{\frac{-1}{2}}H W D_e^{-1} H^TD_v^{\frac{-1}{2}}$ \cite{zhou2006learning}.
		
		\paragraph{Hypergraph Learning}
		
		Hypergraphs are effective for representing complex associations by linking multiple nodes simultaneously, thereby capturing higher-order relationships among vertices \cite{gao2020hypergraph}. Hypergraph learning encompasses the construction and analysis of hypergraphs to represent intricate relationships within data. An essential component of hypergraph learning is the construction of the incidence matrix, which illustrates the connections between vertices and hyperedges \cite{antelmi2023survey}. By learning the incidence matrix, vertex-to-hyperedge connections and hyperedge strengths can be optimized, providing adaptable frameworks for hypergraph construction that capture complex, high-order data relationships. The methods for generating the incidence matrix can be generally divided into two primary categories \cite{zhu2020unsupervised}: fixed and adaptive methods. Fixed methods usually establish the incidence matrix using predetermined criteria or algorithms, and adaptive methods focus on deriving the optimal structure of the hypergraph directly from the data. Each of these categories will be discussed below.

		\begin{enumerate}[leftmargin=5mm, label=(\roman*)]
			\item \textbf{Fixed Methods \cite{zhu2020unsupervised}:}
			Fixed methods for generating incidence matrices in hypergraph learning utilize predetermined algorithms to establish connections between vertices and hyperedges. There are three primary strategies employed in this context:
			
			\begin{itemize}
				
				\item \textbf{Nearest-neighbor-based Strategy \cite{huang2009video}:} It calculates the distances between vertices and uses either the KNN or $\varepsilon$-ball criteria to identify neighbors. Hyperedges are then formed as follows $ e(v) = \mathcal{N}(v) \cup \{v\}$, where \( \mathcal{N}(v) \) represents the neighbors of the vertex \(v\). Consequently, the hyperedge set is defined as $\mathcal{E} = \{e(v) \mid v\in \mathcal{V}\}$.
				
				\item \textbf{Clustering-based Strategy \cite{gao20123}:} It involves grouping vertices using clustering algorithms, such as $k$-means. For \( k \) clusters denoted as \(\mathcal{V}_1, \ldots, \mathcal{V}_k \), hyperedges are constructed as
				$e_l = \mathcal{V}_l$ where $l = 1,\ldots, k$, resulting in the hyperedge set $\mathcal{E} = \{e_1, e_2, \ldots, e_k\}$.
			\end{itemize}

			\item \textbf{Adaptive Methods \cite{zhu2020unsupervised}:}
			Adaptive methods aim to learn key elements of a hypergraph, such as the weight matrix, vertex degree matrix, hyperedge degree matrix, and incidence matrix. These approaches are particularly effective in capturing the complex relationships between vertices and hyperedges, enabling accurate representation of intricate data structures. Among various adaptive methods for learning the incidence matrix, a notable variant is the self-expressive-based method, which typically operates in a two-step process \cite{gao2020hypergraph}. First, it addresses a self-expressive problem; subsequently, it constructs the incidence relationship between hyperedges and their corresponding vertices using the representation coefficient matrix derived from this self-expressive problem.
			
			To illustrate the function of self-expressive-based methods, let us first consider that each data sample \(x_i\), for $i=1,\ldots,n$, represents a vertex within the hypergraph. Additionally, it is assumed that the number of nodes equals the number of hyperedges, meaning \(|\mathcal{V}| = |\mathcal{E}| = n\).
			The three primary variants of self-expressive-based methods can be categorized as follows:
			
			\begin{itemize}
				\item \textbf{$\ell_1$-Hypergraph ($\ell_1$-HG) \cite{7064739}:}
			It integrates a self-expressive problem using the KNN information of each vertex \( v \) within the hypergraph and applies the \( \ell_1 \)-norm to learn a sparse representation coefficient matrix for constructing the hypergraph. The \( \ell_1 \)-HG framework is formulated as follows:
				\begin{align}
					&\min_{z} \|X(v) - Mz\|_2^2 + \alpha \|z\|_1 \\
					&\text{s.t.} \quad z\geq 0,\nonumber
				\end{align}
where \( z \) is the coefficient vector learned for the \( i \)-th nearest neighbor, and \( \alpha \) is the regularization parameter. Moreover, \( X(v) \) denotes the feature vector of the vertex \( v \) within the hypergraph, and \( M \) represents its associated KNN.			
				\item \textbf{$\ell_2$-Hypergraph ($\ell_2$-HG) \cite{jin2019robust}:} It formulates the self-expressive problem by incorporating a locality-preserving constraint on the representation coefficient matrix, enabling the capture of high-order correlations within the data. The $\ell_2$-HG framework is
				\begin{align}
					&\min_{Z} \|X - XZ - E\|_F^2
					+ \alpha\|Z\|_F^2 + \beta\|Q \odot Z\|_F^2
					+ \gamma\|E\|_1\nonumber\\
					&\text{s.t.} \quad Z\in\mathbb{R}^{n\times n}, Z^T\mathbf{1}_n = \mathbf{1}_n, \quad \mathrm{diag}(Z) = 0,\label{19e}
				\end{align}
				where $E$ represents the matrix of data errors, and $Q$ retains the local manifold structures. In \eqref{19e}, $\odot$ denotes the Hadamard product, $\mathbf{1}_n$ is an all-ones vector of size \(n\), and $\alpha, \beta, \gamma$ are the parameters for regularization.
				
				\item \textbf{Elastic-Net Hypergraph (EN-HG) \cite{liu2016elastic}:} It utilizes a combination of the \(\ell_1\) norm and the Frobenius norm to learn a sparse representation coefficient matrix. The EN-HG framework is formulated as follows:
				\begin{align}
					&\min_{Z} \|X - XZ\|_{21} + \|Z\|_1 + \|Z\|_F^2 \\
					&\text{s.t.} \quad Z\in\mathbb{R}^{n\times n}, \mathrm{diag}(Z) = 0.\nonumber
				\end{align}
				
			\end{itemize}
		\end{enumerate}
		
		Finally, self-expressive-based methods construct the incidence relationship between hyperedges and their corresponding vertices using the representation coefficient matrix $Z = [z_{ij}]$ derived from this self-expressive problem. In this case, two strategies are employed as follows \cite{jin2019robust}:
		\begin{itemize}
			\item \textbf{(0-1) Incidence Relationship:}
			This strategy is commonly utilized, where coefficients exceeding a specified threshold are retained. The incidence relation is defined as:
			\begin{equation}
				h(v_i, e_j) =
				\begin{cases}
					1, & \text{if } |z_{ij}| \geq \theta_1 \\
					0, & \text{otherwise}
				\end{cases},
			\end{equation}
			where $i,j=1,\ldots,n$. In this context, the data sample \(x_i\) is linked to the hyperedge \(e_j\) based on the condition that \(z_{ij}\) surpasses the threshold \(\theta_1\).
			
			\item \textbf{Probability Incidence Relationship:}
			In this strategy, the incidence relation is defined through the representation coefficients as follows:
			\begin{equation}
				h(v_i, e_j) =
				\begin{cases}
					|z_{ij}|, & \text{if } |z_{ij}| \geq \theta_2 \\
					0, & \text{otherwise}
				\end{cases},
			\end{equation}
			where $\theta_2$ is a specified threshold.
		\end{itemize}
		
		\paragraph{Matrix Learning Techniques}
		Matrix learning techniques focus on enhancing the hypergraph structural representations through the formulation and solution of an optimization problem. The general framework for a matrix learning technique in hypergraph learning can be outlined as follows \cite{ijcai2017p501}:
		\begin{align}
			&\min_{H, D_e, D_v, W} f(X,H, D_e, D_v, W) + \Omega(W)\\
			&\text{s.t.}\quad W \in \mathcal{C},\nonumber
		\end{align}
		where \( f \) denotes a specific function related to \( (X, H, D_e, D_v, W) \), while \( \Omega(W) \) and \( \mathcal{C} \) denote a regularization term and a set of constraints for \( W \), respectively.
		
		In matrix learning techniques, a common approach for constructing hyperedges is the threshold-based strategy, which allows each sample to have a variable number of nearest neighbors \cite{ijcai2017p501}. Using this approach, the hyperedge \( e(v_j) \) is defined as follows:
		\begin{equation}
			e(v_j) = \{v_i \in \mathcal{V} \mid \theta(g(x_i), g(x_j)) \leq \sigma_j\},
		\end{equation}
		where $i,j=1,\ldots,n$, and \( \sigma_j \) is a predefined threshold. Here, \( \theta(g(x_i), g(x_j)) \) denotes the similarity measure between \( g(x_i) \) and \( g(x_j) \), where \( g \) is a defined mapping applied to the data samples. The resulting hyperedges are then utilized to build the incidence matrix.

		\subsection{Spectral Embedding}
		Spectral embedding is a prominent approach within dimensionality reduction techniques, rooted in the principles of spectral graph theory and manifold learning to uncover fundamental data structures \cite{ghojogh2023elements}. By leveraging the eigenspectrum of graph Laplacian matrices, this method effectively captures complex nonlinear relationships and maintains local geometric characteristics, which are often missed by traditional linear methods. The effectiveness of spectral embedding lies in its ability to reveal the hidden manifold structure through graph spectral analysis, facilitating dimensionality reduction while retaining distinctive features and preserving topological connections within the embedded space \cite{ding2024survey}.
		
		Given a similarity graph that encodes the pairwise relationships between data points \(\{x_1, \ldots, x_n\}\), the objective for spectral embedding optimization can be formulated as:
		\begin{align}
			&\min_{F} Tr(F^T L F) \\
			&\text{s.t.} \quad F \in \mathbb{R}^{n \times k},\, F^T F = I_k,\nonumber
		\end{align}
		where \( F \) denotes the spectral embedding matrix, and \( k \) represents the number of dimensions into which the original data is embedded. Moreover, the graph Laplacian matrix \( L \) can be constructed in three principal forms \cite{von2007tutorial}:
		\begin{itemize}
			\item\textbf{Unormalized Graph Laplacian Matrix:} \[L = D - W.\]
			\item\textbf{Normalized Graph Laplacian Matrix (Symmetric):} \[L_{sym} = I_n - D^{-\frac{1}{2}}WD^{-\frac{1}{2}}.\]
			\item\textbf{Normalized Graph Laplacian Matrix (Random Walk):} \[L_{rw} = I_n - D^{-1}W.\]
		\end{itemize}
		It should be highlighted that the spectrum of the graph Laplacian matrix \( L \) reveals information about the connectivity of the graph. The orthogonality condition \( F^T F = I_k \) in spectral embedding ensures that the embedding space is constructed from the eigenvectors corresponding to the smallest non-zero eigenvalues of \( L \). Selecting these eigenvectors not only preserves the connectivity within clusters but also provides a low-dimensional representation that reflects the cluster structure of the graph.

		\subsection{Partitioning Process}
		In spectral clustering, the partitioning process assigns cluster memberships by leveraging a low-dimensional spectral representation obtained through spectral embedding. Partitioning approaches are classified into two main types--two-step and one-step clustering--depending on how spectral embedding and clustering are conducted \cite{YUN2023137}.
		
		\subsubsection{Two-Step Clustering}
		In conventional spectral clustering approaches, the clustering task is generally conducted in two primary steps. Initially, these methods relax the discrete constraints and apply eigenvalue decomposition to the graph Laplacian matrix, resulting in a spectral embedding matrix \(F\) in a continuous form. Next, a clustering method such as K-means is performed on \(F\) to transform these continuous embeddings into discrete cluster assignments.
		
		While two-step clustering presents a straightforward framework, the resulting optimal solution \(F\) may not closely resemble the true discrete clustering solution, as spectral embeddings do not inherently exhibit the required discrete structure.
		Additionally, K-means is sensitive to its initial settings and often converges to locally optimal solutions. Thus, using the obtained embedding matrix \(F\) as the basis for clustering may introduce inaccuracies, potentially compromising the clustering results.

		\subsubsection{One-Step Clustering}
		One-step spectral clustering algorithms utilize a unified framework that combines the learning of spectral embeddings with cluster assignments into a single process. This simultaneous process allows for the direct calculation of the final clustering outcome, improving both efficiency and effectiveness in comparison to two-step clustering methods. In this context, one-step spectral clustering techniques can be divided into two primary categories: continuous and discrete frameworks.
		
		\paragraph{Continuous Framework}
		
		This type of one-step spectral clustering typically learns the embedding matrix so that the data points can be divided into \( k \) clusters without the need for any discretization steps. The two primary methods based on the continuous framework are as follows:
		
		\begin{enumerate}[leftmargin=5mm, label=(\roman*)] 
			
			\item \textbf{Laplacian-Based Methods:}
			These methods aim to learn the Laplacian matrix of a similarity graph consisting of \( k \) connected components. To achieve this, they leverage the Ky Fan's theorem alongside a fundamental principle in graph learning \cite{von2007tutorial}: the multiplicity \( k \) of the eigenvalue 0 of the graph Laplacian matrix \( L \) indicates the number of connected components in the similarity graph. This relationship emphasizes the critical role of the eigenvalues of the graph Laplacian in understanding the connectivity of graph structures, facilitating more effective clustering and analysis of complex data. Inspired by this insight, a notable example of Laplacian-based methods is the CAN approach \cite{nie2014clustering}, whose objective function was described in Problem \eqref{can1} as an adaptive neighbor method:
			\begin{align}
				&\min_{s_{ij},F} \sum_{i,j=1}^n (\Vert x_i - x_j\Vert_2^2 s_{ij} + \gamma s_{ij}^2) + \nu Tr(F^T L_S F) \label{can1one}\\
				&\text{s.t.}\quad\sum_{j=1}^n s_{ij} =1, 0 \leq s_{ij} \leq 1,\,F\in \mathbb{R}^{n \times k},\, F^T F = I_k,\nonumber
			\end{align}
			where \( \nu \) is the regularization parameter. Here, to facilitate the learning of an optimal neighbor assignment matrix \( S \) with a distinct clustering structure, the following problem, derived from the Ky Fan's theorem for the Laplacian matrix \( L_S \), is defined as:
			\begin{align}
				\sum_{i=1}^k\lambda_i(L_s) = \min_{F\in \mathbb{R}^{n \times k},\, F^T F = I_k} Tr(F^T L_S F),
			\end{align}
			where \(L_S \) is associated with \( (S + S^T)/2 \), and $\lambda_i(L_S)$ denotes the $i$-th eigenvalue of \(L_S\), for $i=1,\dots,k$.
			
			\item \textbf{Matrix Factorization-Based Methods:}
			These methods take advantage of matrix factorization techniques for their time efficiency while also utilizing spectral clustering properties to uncover non-linear structural information. A key example is the Nonnegative Embedding and Spectral Embedding (NESE) method \cite{HU2020251}, which is rooted in the relationship between symmetric nonnegative matrix factorization and spectral clustering \cite{kuang2015symnmf}. The optimization problem for NESE is defined as follows:
			\begin{align}
				&\min_{F,H} \|D^{-\frac{1}{2}} WD^{-\frac{1}{2}} - HF^T\|^2_F \label{can10one}\\
				&\text{s.t.}\quad F, H \in \mathbb{R}^{n \times k}, \, F^T F = I_k,\, H\geq 0,\nonumber
			\end{align}
			where \( F \) represents the spectral embedding, and \( H \) denotes the nonnegative embedding matrix that categorizes each \( i \)-th data sample into the class indicated by the position of the highest value in the \( i \)-th row of \( H \).
		\end{enumerate}
		
		\paragraph{Discrete Framework}
		This one-step spectral clustering method basically utilizes three techniques to learn the embedding matrix: Spectral Rotation (SR) \cite{huang2013spectral}, Discrete Spectral Clustering (DSC) \cite{luo2019discrete}, and Discrete Nonnegative Spectral Clustering (DNSC) \cite{yang2017discrete}.
		
		\begin{enumerate}[leftmargin=5mm, label=(\roman*)] 
			
		\item \textbf{SR-Based Methods:}
		SR \cite{huang2013spectral} is a method designed to transform a continuous spectral embedding matrix, typically obtained from spectral clustering, into a binary matrix that more accurately represents discrete cluster indicators. Given a spectral embedding matrix \( F \), SR seeks to determine an optimal orthogonal transformation that maps the real-valued matrix \( F \) to a binary cluster indicator matrix $Y$. The optimization problem for SR can be formulated as:
		\begin{align}
			&\min_{Y, R} \|F-YR^T\|^2_F \label{can100one}\\
			&\text{s.t.}\quad R \in \mathbb{R}^{k \times k},\,Y\in \text{Ind},\, \, R^T R = I_k,\nonumber
		\end{align}
		where \( Y \in \text{Ind} \) denotes the clustering outcome as a binary indicator matrix. Following this description, a basic outline for using spectral embedding and SR together is \cite{pang2018spectral}:
		\begin{align}
			&\min_{F, Y, R} Tr(F^T L_{sym} F) + \|F-YR^T\|^2_F \label{can1000one}\\
			&\text{s.t.}\, F\in \mathbb{R}^{n \times k}, R \in \mathbb{R}^{k \times k}, F^T F = I_k, Y\in \text{Ind}, R^T R = I_k.\nonumber
		\end{align}
		
		\item \textbf{DSC-Based Methods:}
These methods adopt a direct framework to compute a discrete-valued spectral embedding matrix by solving the following optimization problem:
				\begin{align}
					&\min_{F,Y} Tr(F^T L F) - \nu Tr(F^T Y) \\
					&\text{s.t.} \quad F \in \mathbb{R}^{n \times k},\, F^T F = I_k,\, Y\in \text{Ind},\nonumber
				\end{align}
where \(\nu\) denotes a regularization parameter. Notably, for any \(Y \in \text{Ind}\), the optimal solution for the spectral embedding matrix is given by \(F = Y (Y^T Y)^{-\frac{1}{2}}\). This approach directly derives the discrete embedding matrix \(F\) without relying on conventional relax-and-discretize methods.

		\item \textbf{DNSC-Based Methods:}
		DNSC \cite{yang2017discrete} is an alternative method for generating a discrete-valued spectral embedding matrix. In the absence of sign constraints on the spectral embedding matrix \( F \), it may include both positive and negative values, deviating from the structure of a true cluster indicator matrix, where each row has exactly one positive element, with all others being zero. To resolve this, DNSC enforces both nonnegativity and orthogonality constraints on \( F \), resulting in the following problem:
		\begin{align}\label{dnsc1}
			&\min_{F} \text{Tr}(F^T L F) \\
			&\text{s.t.} \quad F \in \mathbb{R}^{n \times k},\, F^T F = I_k,\, F\geq 0.\nonumber
		\end{align}
Essentially, by introducing a nonnegative constraint, the DNSC problem \eqref{dnsc1} becomes more consistent with the N-cut method described in \eqref{dnsc0} for producing discrete indicator vectors \cite{ijcai2017p251}. This improves the interpretability of the spectral embedding matrix \( F \) and eliminates the need for post-processing techniques like k-means.
	\end{enumerate}
	 	
\section{Category of Spectral Clustering} \label{category}
Spectral clustering methods are systematically categorized into single-view and multi-view approaches. Single-view spectral clustering constructs graph representations from individual data perspectives, whereas multi-view approaches integrate information from multiple data representations to achieve enhanced clustering performance. Both paradigms share three fundamental stages: graph structure learning, spectral embedding, and partition process. Multi-view spectral clustering extends this framework by incorporating two additional components: view-specific matrix construction and cross-view information fusion. This extended framework enables the discovery of intrinsic data patterns through complementary view relationships that remain undetectable in single-view analysis, leading to more comprehensive clustering solutions.

\subsection{Single-view Spectral Clustering}
Single-view spectral clustering encompasses diverse methodologies for graph-based data partitioning, distinguished by graph structure learning approaches and partitioning processes. This section presents a systematic analysis through two fundamental perspectives: graph structure paradigms and partitioning strategies. The methodology spans three principal architectures: pairwise graphs, anchor graphs, and hypergraphs. Within each, methods are categorized as fixed approaches using predetermined rules or adaptive techniques dynamically learning optimal structures. The partitioning process employs either a two-step approach with sequential spectral embedding and clustering, or a one-step approach integrating these processes into a unified optimization framework.

\subsubsection{Pairwise Graph}
Pairwise graphs represent relationships between data points and form the foundation of many spectral clustering algorithms. These graphs can be constructed using fixed or adaptive approaches, offering varying levels of flexibility and performance in capturing the underlying structure of the data.

\textbf{Fixed Methods:} Fixed methods establish graph topology through predetermined protocols, forming the foundational approach in graph construction. These methods, detailed in Table \ref{tab3}, encompass KNN, mutual KNN, fully connected graphs, and $\epsilon$-ball graphs. Seminal works include von Luxburg's mutual kNN formulation \cite{von2007tutorial}, which established theoretical foundations, and subsequent enhancements such as KNN-SC \cite{kim2021knn} for adaptive neighbor selection. Notable advancements include SR \cite{huang2013spectral} for improved discretization and JSESR \cite{pang2018spectral}, which integrates spectral embedding and rotation in a unified framework. These approaches offer computational efficiency while maintaining consistent structural properties across diverse datasets.

\textbf{Adaptive Neighbor Methods:}\textbf{Adaptive Neighbor Methods:} 

Adaptive Neighbor Methods focus on dynamically constructing graph structures directly from the relationships within the data, tailoring the clustering process to the dataset's underlying structure and enhancing the flexibility of clustering algorithms. CAN \cite{nie2014clustering} pioneered this methodology by learning probabilistic neighbor connections during clustering, ensuring that the graph representation captures meaningful local relationships and global patterns in the data. PCAN \cite{nie2014clustering} extended this framework to accommodate high-dimensional datasets by jointly optimizing dimensionality reduction and similarity learning, creating a unified model that improves clustering quality while reducing computational complexity. SWCAN \cite{nie2020self} introduced adaptive feature weighting, enabling the model to prioritize features with higher relevance to clustering tasks, effectively mitigating the influence of noisy or redundant features. ERCAN \cite{wang2022entropy} further refined these techniques by incorporating entropy regularization, which not only optimizes the similarity distribution but also balances the adaptiveness of neighbor assignment with robust clustering objectives, leading to improved performance on diverse datasets.

SR-OSC \cite{zhu2020spectral} took a step further by unifying graph construction with clustering optimization in a low-dimensional space, leveraging spectral rotation to enhance stability and accuracy. SCANDLE \cite{zhao2021spectral} revolutionized the field by integrating deep learning techniques, replacing traditional eigen-decomposition with neural networks, thereby improving scalability and effectiveness for large and complex datasets. PTAG \cite{zhou2023typicality} introduced a groundbreaking typicality-based neighbor assignment framework, which eliminates the conventional sum-to-one constraints in similarity assignment and incorporates block-diagonal regularization. This innovation strengthens cluster separability and provides increased robustness against noise and outliers. These advancements, summarized in Table~\ref{tab3}, underscore the evolution of spectral clustering toward adaptive and efficient single-step frameworks capable of handling large- and high-dimensional data while maintaining precision and robustness.

\textbf{Initial Similarity Matrix Methods:} Initial similarity matrix methods refine pre-constructed graphs to improve clustering performance. These approaches have gained prominence for their ability to directly optimize graph structures for clustering tasks. The development of these methods can be traced through several key algorithms. The Constrained Laplacian Rank (CLR) algorithm \cite{nie2016constrained} learns a new similarity matrix with exactly $k$ connected components from an initial similarity matrix, employing a one-step continuous approach through Ky Fan theory to reveal cluster structure without post-processing.  Building upon CLR, RCFE \cite{li2018rank} enhances spectral clustering by simultaneously learning adaptive graph structure and low-dimensional embedding, using Laplacian rank constraints to recover block-diagonal similarity matrices. For high-dimensional data, RCFE implements a flexible embedding scheme allowing mismatch between projected and embedded features. SClump \cite{li2019spectral} addresses heterogeneous information networks by integrating multiple meta-path similarities, optimizing similarity matrices and meta-path weights simultaneously. The method employs} Laplacian rank constraints with iterative refinement. More recently, FCLR \cite{duan2024new} accelerated CLR using anchors and bipartite graph learning strategies, reducing computational complexity from quadratic to linear with respect to data size.

\begin{table*}[!h]
  \centering
  \caption{Comparison of Fixed, Adaptive Neighbor, and Initial Similarity Matrix Methods for Spectral Clustering}
  \label{tab3}
  \scriptsize 
  \setlength{\tabcolsep}{1pt} 
  \renewcommand{\arraystretch}{1} 
  \begin{tabular}{|c|c|c|c|c|c|}
 \hline
        \multirow{3}{*}{\#} & \multirow{3}{*}{\textbf{Method}} & \multirow{3}{*}{\textbf{Graph Structure Learning}} & \multicolumn{3}{c|}{\textbf{Partitioning Process}} \\ \cline{4-6}
        & & & \multirow{2}{*}{\textbf{Two-Step}} & \multicolumn{2}{c|}{\textbf{One-Step}} \\ \cline{5-6}
        & & & & \textbf{Continuous} & \textbf{Discretized} \\ \hline
    1 & MKNNG (2007) \cite{von2007tutorial} & Mutual-KNN & \checkmark & & \\
    \hline
    2 & SRC (2013) \cite{huang2013spectral} & KNN & & & \checkmark \\
    \hline
    3 & JSESR (2020) \cite{pang2018spectral} & KNN & & & \checkmark \\
    \hline
    4 & KNN-SC (2021) \cite{kim2021knn} & KNN & \checkmark & & \\
    \hline
    5 & CAN (2014) \cite{nie2014clustering} & $\min\limits_{s_{ij}} \sum_{i,j=1}^n (\Vert x_i - x_j\Vert_2^2 s_{ij} + \gamma s_{ij}^2)$ & & \checkmark & \\
    \hline
    6 & PCAN (2014) \cite{nie2014clustering} & $\min\limits_{s_{ij},P} \sum_{i,j=1}^n (\Vert P^T x_i - P^T x_j\Vert_2^2 s_{ij} + \gamma s_{ij}^2)$ & & \checkmark & \\
    \hline
    7 & SEANC (2019) \cite{wang2018spectral}   & $\min\limits_{s_{ij}} \sum_{i,j=1}^n (\Vert f_i - f_j\Vert_2^2 s_{ij} + \gamma s_{ij}^2 + \beta \Vert g_i - g_j\Vert_2^2 s_{ij})$ &  & \checkmark  & \\
    \hline
    8 & SR-OSC (2020) \cite{zhu2020spectral} & $\min\limits_{s_{ij},P} \sum_{i,j=1}^n (\Vert P^T x_i - P^T x_j\Vert_2^2 s_{ij} + \gamma s_{ij}^2)$ & & & \checkmark \\
    \hline
    9 & SWCAN (2020) \cite{nie2020self}   & $\min\limits_{s_{ij},\Theta} \sum_{i,j=1}^n (\Vert \Theta x_i - \Theta x_j\Vert_2^2 s_{ij} + \gamma s_{ij}^2)$
     & & \checkmark & \\
    \hline
    10 & ERCAN (2022) \cite{wang2022entropy} & 
        $\min\limits_{s_{ij}} \sum_{i,j=1}^n (\Vert x_i - x_j\Vert_2^2 s_{ij} + \gamma s_{ij}\log s_{ij})$
   & & \checkmark & \\
    \hline
    11 & SCANDLE (2023) \cite{zhao2021spectral}  & $\min\limits_{s_{ij}} \sum_{i,j=1}^n (\Vert x_i - x_j\Vert_2^2 s_{ij} + \gamma s_{ij}^2)$ & \checkmark &  & \\
    \hline
    12 &  CTAG (2023) \cite{zhou2023typicality} & $\min\limits_{s_{ij}} \sum_{i,j=1}^n \Vert x_i - x_j\Vert_2^2 s_{ij} + \sum_{i=1}^n\gamma_i \sum_{j=1}^n (s_{ij}\log s_{ij}-s_{ij})$   
        & & \checkmark & \\
    \hline
    13 & CLR (2016) \cite{nie2016constrained} & $\min\limits_{S} \|S - A\|_F^2$ & & \checkmark & \\
    \hline
    14 & RCFE (2018) \cite{li2018rank} & $\min\limits_{S} \|S - A\|_F^2 + \alpha \|X^T P - F\|_F^2$ & & \checkmark & \\
    \hline
    15 & SClump (2019) \cite{li2019spectral} & $\min\limits_{S} \| S - A \|_F^2 + \alpha\| S \|_F^2$ & & \checkmark & \\
    \hline
    16 & FCLR (2024) \cite{duan2024new} & $\min\limits_{S} \| S - A \|_F^2 $ & & \checkmark & \\
    \hline
  \end{tabular}
\end{table*}

\textbf{Self-expressive Methods:} Self-expressive methods construct similarity matrices by representing each data point as a linear combination of others in the dataset. This results in a coefficient matrix that encodes intrinsic relationships and subspace structures, effectively capturing the underlying data distribution.

Sparse Subspace Clustering (SSC) \cite{elhamifar2013sparse} pioneered the use of $\ell_1$-norm minimization to learn a sparse coefficient matrix. This method is particularly effective in dealing with data derived from multiple linear subspaces by ensuring that each data point is only connected to points within its own subspace, achieving high precision in clustering. Low-Rank Representation (LRR) \cite{liu2012robust} extended this framework by employing nuclear norm minimization to obtain a low-rank representation. This approach excels at capturing global structural information while demonstrating robustness to noise and outliers, making it suitable for diverse real-world applications. Least Squares Regression (LSR) \cite{lu2012robust} introduced Frobenius norm minimization, which simplified the optimization process through a closed-form solution. This not only reduced computational complexity but also maintained high efficiency, making it a practical choice for large-scale datasets.

Smooth Representation (SMR) \cite{hu2014smooth} enhanced clustering performance by incorporating a smoothness term derived from a $k$-nearest neighbor graph. This innovation preserved local structural information while enforcing a grouping effect among similar data points, ensuring better consistency in the generated clusters. Laplacian Regularized LRR (LapLRR) \cite{liu2014enhancing} integrated manifold regularization into the low-rank representation framework. By simultaneously preserving global Euclidean structure and local manifold properties, this method effectively balanced global and local data characteristics for improved clustering accuracy. Dual Graph Regularized LRR (DGLRR) \cite{yin2015dual} further advanced this idea by introducing dual graph regularization to model both data and feature manifolds. This dual perspective captured a more comprehensive view of the relationships within the data, improving clustering performance on complex datasets.

Recent advancements have focused on leveraging deep learning and adaptive techniques to enhance clustering. Graph Compact LRR (GCLRR) \cite{du2017graph} combined dictionary learning with clustering, modeling the dictionary as a linear combination of data points to achieve compact and efficient representations. Deep Subspace Clustering Networks (DSC-Nets) \cite{ji2017deep} utilized self-expressive layers within autoencoder architectures, enabling the learning of nonlinear data mappings and achieving significant improvements in clustering for non-linear subspaces. Similarly, Structured AutoEncoders (StructAE) \cite{peng2018structured} incorporated self-expressive layers while adding structural constraints, further boosting clustering accuracy for complex datasets. LRR with Adaptive Graph Regularization (LRR-AGR) \cite{wen2018low} improved graph construction by incorporating distance regularization and enforcing non-negative constraints, adapting dynamically to the data for better clustering results. 

Low-Rank Sparse Subspace (LSS) \cite{zhu2018low} proposed a dynamic approach for learning similarity matrices directly from low-dimensional spaces, enhancing its ability to identify intrinsic structures within the data. Subspace Clustering via Iterative Discrete Optimization (SSC-ID) \cite{tong2020one} introduced a robust one-step clustering method that employed spectral rotation and self-paced learning, ensuring resilience against noisy data and enhancing clustering quality. The latest development, Subspace Clustering via Joint $\ell_{1,2}$ and $\ell_{2,1}$ norms (SCJL12-21) \cite{dong2022subspace}, tackled the challenge of heavily corrupted data by applying joint norm minimization, effectively excluding corrupted data points while maintaining robustness to outliers.

Table~\ref{tab:self_expressive_methods} provides a detailed comparison of these methods, outlining their graph construction strategies and clustering steps.

\begin{table*}[!h]
  \centering
    \caption{Comparison of Self-Expressive Methods Spectral Clustering Methods: Graph Construction and Clustering Steps}
  \label{tab:self_expressive_methods}
  \scriptsize 
  \setlength{\tabcolsep}{1pt} 
  \renewcommand{\arraystretch}{1} 
  \begin{tabular}{|c|c|c|c|c|c|c|}
        \hline
        \multirow{3}{*}{\#} & \multirow{3}{*}{\textbf{Method}} & \multicolumn{2}{c|}{\multirow{2}{*}{\textbf{Graph Structure Learning}}} & \multicolumn{3}{c|}{\textbf{Partitioning Process}} \\ \cline{5-7}
    &    &    \multicolumn{1}{c}{}     &    \multicolumn{1}{c|}{} & \multirow{2}{*}{\textbf{Two-Step}} & \multicolumn{2}{c|}{\textbf{One-Step}} \\ \cline{3-4}\cline{6-7}
    &    & \multicolumn{1}{c|}{$\textbf{Loss}(X,Z)$} & \multicolumn{1}{c|}{\textbf{Regularization}} & & \textbf{Continuous} & \textbf{Discretized} \\ \hline
    1 & SSC (2012) \cite{elhamifar2013sparse} &  \multicolumn{1}{c|}{$\|X - XZ\|_1$} &  \multicolumn{1}{c|}{$\|Z\|_1$} & $\checkmark$ & & \\
    \hline
    2 & LRR (2013) \cite{liu2012robust} & $\|X - XZ\|_{21}$ & $\|Z\|_*$ & $\checkmark$ & & \\
    \hline
    3 & LSR (2013) \cite{lu2012robust} & $\|X - XZ\|_F$ & $\|Z\|_F^2$ & $\checkmark$ & & \\
    \hline
    4 & SMR (2014) \cite{ hu2014smooth} & $\|X - XZ\|_{21}$ & $\text{tr}(Z L Z^T)$ & $\checkmark$ & & \\
    \hline
    5 & LapLRR (2014) \cite{liu2014enhancing}  & $\|X - XZ\|_F$ & $\|Z\|_* + \text{tr}(Z L Z^T)$ & $\checkmark$ & & \\
    \hline
    6 & DGLRR (2015) \cite{yin2015dual} & $\|X - XZ - GX\|_1$ & $\begin{array}{c} \|Z\|_* + \|G\|_* + \text{tr}(Z L_Z Z^T) + \text{tr}(G L_G G^T) \end{array}$ & $\checkmark$ & & \\
    \hline
    7 & GCLRR (2017) \cite{du2017graph} & $\|X - XZ\|_{21}$ & $\|Z\|_* + \text{tr}(Z L Z^T)$ & $\checkmark$ & & \\
    \hline
   8 & DSC-Nets (2017) \cite{ji2017deep} & $\|X - \hat{X}\|_F^2 + \|H - HZ\|_F^2$ & $\|Z\|_p$ & \checkmark & & \\
    \hline
    9 & StructAE (2018) \cite{peng2018structured}  & $\|X - H^{(M)}\|_F^2 + \|H^{(M/2)} - H^{(M/2)}Z\|_F^2$ &  $\sum_{m=1}^M(\| W^{(m)} \|_F^2 + \| b^{(m)} \|_2^2)$ & $\checkmark$ & & \\
    \hline
    10 & LRR-AGR (2018) \cite{wen2018low} & $\|X - XZ\|_1$ & $\|Z\|_*$ & & $\checkmark$ & \\
    \hline
    11 & LSS (2019) \cite{zhu2018low}  & $\|P^TX - P^TXZ^{(1)}\|_F^2+\|X - XZ^{(2)}\|_F^2$ & $\sum_{i=1}^n (\Vert Z_i^{(1)} - Z_i^{(2)}\Vert_2^2
    +  \| P \|_{21}$ & & $\checkmark$ & \\
    \hline
    12 & SSC-ID (2020) \cite{tong2020one} & $\|X - XZ\|_F^2$ & $\|Z\|_1$ & & & $\checkmark$ \\
    \hline
    13 & SCJL12-21 (2024) \cite{dong2022subspace} & $\|X - XZ\|_{21}$ & $\|Z\|_{12}$ & $\checkmark$ & & \\
    \hline
 \end{tabular}
\end{table*}

\subsubsection{Anchor Graph}

Anchor Graph methods have significantly enhanced spectral clustering for large-scale data by replacing the full similarity matrix with a more efficient representation based on anchor points. This innovation reduces computational complexity from $O(n^2)$ or $O(n^3)$ to $O(nm)$, where $n$ represents the number of data points and $m$ is the number of anchors ($m \ll n$). 

LSC \cite{cai2014large} introduced $k$-means clustering for anchor selection and employed a fixed Gaussian kernel for graph construction, following a two-step clustering process. FSC \cite{zhu2017fast} advanced the field by adopting BKHK for anchor selection and introducing an adaptive, parameter-free approach for graph construction, while retaining the two-step process. FSCAG \cite{wang2017fast} utilized random anchor selection combined with adaptive graph construction and transitioned to a continuous one-step clustering framework. SGCNR \cite{wang2019scalable} applied $k$-means for anchor selection alongside adaptive graph construction and incorporated nonnegative relaxation within a continuous one-step clustering model. RLGCH \cite{yang2022robust} implemented $k$-means++ for anchor selection and a fixed Gaussian kernel, employing a discretized one-step clustering method enhanced by the $\ell_{2,1}$-norm to improve robustness.

FGC-SS \cite{chen2022fgc_ss} utilized BKHK for anchor selection with adaptive graph construction, unifying spectral embedding and rotation within a discretized one-step framework. EDCAG \cite{wang2023efficient} introduced BKHK for anchor selection and adaptive graph construction while leveraging a bipartite similarity matrix in a discretized one-step clustering process. RESKM \cite{yang2023reskm} reverted to $k$-means for anchor selection and employed fixed graph construction, maintaining the traditional two-step clustering process. FSBGL \cite{yang2023fast} employed $k$-means clustering for anchor selection, combined with adaptive graph construction, and utilized a bipartite similarity matrix within a continuous one-step framework. SE-ISR \cite{wang2021fast} also adopted $k$-means for anchor selection, integrating adaptive graph construction and unifying spectral embedding with rotation in a discretized one-step process. SCLE \cite{gao2024spectral} applied BKHK for anchor selection and adaptive graph construction, introducing linear embedding into a discretized one-step clustering framework.

\begin{table*}[!h]
  \centering
  \caption{Comparison of Anchor Graph Methods Spectral Clustering Methods: Graph Construction and Clustering Steps}
  \label{tab5}
  \scriptsize 
  \setlength{\tabcolsep}{1pt} 
  \renewcommand{\arraystretch}{1} 
  \begin{tabular}{|c|c|c|c|c|c|c|c|}
        \hline
        \multirow{3}{*}{\#} & \multirow{3}{*}{\textbf{Method}} & \multicolumn{3}{c|}{\multirow{2}{*}{\textbf{Graph Structure Learning}}} & \multicolumn{3}{c|}{\textbf{Partitioning Process}} \\ \cline{6-8}
    &   &    \multicolumn{1}{c}{}  &    \multicolumn{1}{c}{}     &    \multicolumn{1}{c|}{} & \multirow{2}{*}{\textbf{Two-Step}} & \multicolumn{2}{c|}{\textbf{One-Step}} \\ \cline{3-5}\cline{7-8}
    &    & \multicolumn{1}{c|}{\textbf{Select Anchor}} & \multicolumn{1}{c|}{\textbf{Anchor Graph}} & \multicolumn{1}{c|}{\textbf{Similarity Matrix}} & & \textbf{Continuous} & \textbf{Discretized} \\ \hline
    1 & LSC (2015) \cite{cai2014large} & K-means & Fixed & Gram & \checkmark &  & \\
    \hline
    2 & FSC (2017) \cite{zhu2017fast} & BKHK & Adaptive & Gram & \checkmark &  & \\
    \hline
    3 & FSCAG (2017) \cite{wang2017fast} & Random & Adaptive & Gram &  & \checkmark & \\
    \hline 
    4 & SGCNR (2019) \cite{wang2019scalable} & K-means & Adaptive & Gram &  & \checkmark &  \\
    \hline
    5 & RLGCH (2022) \cite{yang2022robust} & K-means++ & Fixed & Gram &  &  & \checkmark \\
    \hline
    6 & FGC-SS (2022) \cite{chen2022fgc_ss} & BKHK & Adaptive & Gram &  &  & \checkmark \\
    \hline
    7 & EDCAG (2023) \cite{wang2023efficient} & BKHK & Adaptive & Bipartite &  &  & \checkmark \\
    \hline 
    8 & RESKM (2023) \cite{yang2023reskm} & K-means & Fixed & Gram & \checkmark &  & \\
    \hline
    9 & FSBGL (2023) \cite{yang2023fast} & K-means & Adaptive & Bipartite &  & \checkmark & \\
    \hline
    10 & SE-ISR (2023) \cite{wang2021fast} & K-means & Adaptive & Gram &  &  & \checkmark \\
    \hline
    11 & SCLE (2024) \cite{gao2024spectral} & BKHK & Adaptive & Gram &  &  & \checkmark \\
    \hline
\end{tabular}
\end{table*}

\subsubsection{Hypergraph Methods}

Hypergraph spectral clustering methods encompass three main approaches: hypergraph Laplacians \cite{gao2020hypergraph}, hypergraph reduction \cite{huang2017effect}, and tensor modeling \cite{ghoshdastidar2015provable}. Hypergraph Laplacian methods generalize graph Laplacians to higher-order data structures, preserving the original hypergraph topology. Hypergraph reduction transforms hypergraphs into standard graphs through methods such as Star-Expansion \cite{pu2012hypergraph}, Clique-Expansion \cite{saito2018hypergraph}, and Line-Expansion \cite{yang2020hypergraph}. Tensor modeling represents uniform hypergraphs as higher-dimensional tensors \cite{ghoshdastidar2015provable}. This survey primarily focuses on hypergraph Laplacian methods, which yield accurate results for complex, high-order relationships while avoiding information loss associated with reduction techniques. These methods effectively capture intrinsic multi-way relationships in real-world datasets by operating directly on the hypergraph structure, although they come with higher computational costs.

Hypergraph methods can be broadly categorized into fixed and adaptive construction approaches. Among fixed construction methods, Normalized Hypergraph Spectral Clustering (NHSC) \cite{zhou2006learning} extends spectral clustering to hypergraphs using k-nearest neighbor (KNN) hypergraph construction. Unsupervised Image Categorization (UIC) \cite{huang2011unsupervised} also employs KNN-based hypergraph construction for image analysis. Context-Aware Hypergraph Spectral Clustering (CAHSM) \cite{li2013context} combines multiple similarity measures within a fixed hypergraph structure to enhance clustering results. Hypergraph Spectral Clustering with Local Refinement (HSCLR) \cite{ahn2018hypergraph} uses a k-means-based hypergraph construction for improved accuracy. More recent fixed approaches include Hypergraph Spectral Clustering (HSC) for point cloud segmentation \cite{zhang2020hypergraph} and GraphLSHC \cite{yang2021graphlshc}, both of which utilize KNN-based hypergraph construction for efficient processing.

Adaptive construction methods provide more flexibility in capturing data relationships, allowing them to better handle complex patterns. The Regression-based Hypergraph (RH) \cite{huang2016regression} method employs sparse representation to construct hypergraphs, facilitating more precise clustering. Elastic-Net Hypergraph (ENHG) \cite{liu2016elastic} uses elastic net regularization to improve the stability and accuracy of the constructed hypergraph. Correntropy-Induced Robust Low-Rank Hypergraph (CIRLH) \cite{jin2018correntropy} combines low-rank representation with a correntropy-induced metric, providing robustness to noise. The Robust $\ell_2$-Hypergraph (RH$\ell_2$) \cite{jin2019robust} employs affine subspace ridge regression for better clustering performance under noisy conditions. More recent adaptive methods, such as Elastic-Net constrained Hypergraph (EN-cHG) \cite{jin2020link} and Hessian Non-negative Hypergraph (HNH) \cite{li2023hessian}, integrate various regularization techniques to enhance clustering performance and improve robustness against high-dimensional data and noise.

Table \ref{tab:self_expressive_methodshyper} provides a comprehensive comparison of these hypergraph-based spectral clustering methods, highlighting their graph construction approaches and clustering steps. As shown in the table, all methods employ a two-step clustering approach, regardless of the construction method used. The transition from fixed to adaptive construction techniques reflects the ongoing effort to enhance hypergraph learning’s ability to capture complex data relationships, address challenges like noise, and handle high-dimensional data, thereby improving clustering performance in real-world applications.

\begin{table}[!h]
  \centering
  \caption{Comparison of Hypergraph-Based Spectral Clustering Methods: Graph Construction and Clustering Steps}
  \label{tab:self_expressive_methodshyper}
  \scriptsize 
  \setlength{\tabcolsep}{1pt} 
  \renewcommand{\arraystretch}{1} 
  \scalebox{0.9}{\begin{tabular}{|c|c|c|c|c|c|}
 \hline
        \multirow{3}{*}{\#} & \multirow{3}{*}{\textbf{Method}} & \multirow{3}{*}{\textbf{Graph Structure Learning}} & \multicolumn{3}{c|}{\textbf{Partitioning Process}} \\ \cline{4-6}
        & & & \multirow{2}{*}{\textbf{Two-Step}} & \multicolumn{2}{c|}{\textbf{One-Step}} \\ \cline{5-6}
        & & & & \textbf{Continuous} & \textbf{Discretized} \\ \hline
    1 & NHSC (2006) \cite{zhou2006learning} & KNN-HG & \checkmark & & \\
    \hline 
    2 & UIC (2011) \cite{huang2011unsupervised} & KNN-HG & \checkmark & & \\
    \hline 
    3 & CAHSM (2014) \cite{li2013context} & KNN-HG & \checkmark & & \\
    \hline
    4 & HSCLR (2018) \cite{ahn2018hypergraph} & K-means-HG & \checkmark & & \\
    \hline
    5 & HSC (2020) \cite{zhang2020hypergraph} & KNN-HG & \checkmark & & \\
    \hline
    6 & GLSHC (2021) \cite{yang2021graphlshc} & KNN-HG & \checkmark & & \\
    \hline 
    7 & RH (2016) \cite{huang2016regression} & Sparse Representation & \checkmark & & \\
    \hline  
    8 & ENHG (2017) \cite{liu2016elastic} & EN-HG & \checkmark & & \\
    \hline
    9 & CIRLH (2019) \cite{jin2018correntropy} & Correntropy-Induced & \checkmark & & \\
    \hline
    10 & RH$L_2$ (2019) \cite{jin2019robust} & L2-HG & \checkmark & & \\
    \hline 
    11 & EN-cHG (2020) \cite{jin2020link} & EN-HG & \checkmark & & \\
    \hline
    12 & HNH (2023) \cite{li2023hessian} & HN-HG & \checkmark & & \\
    \hline
  \end{tabular}}
\end{table}

 \subsection{Multi-view Spectral Clustering}
Multi-view spectral clustering represents a significant advancement over traditional spectral clustering by incorporating multiple complementary representations of data \cite{fang2023comprehensive}. This methodology addresses scenarios where data exhibits heterogeneous characteristics captured through diverse feature sets, measurement modalities, or representation schemes \cite{li2018survey}. For example, multimedia content analysis often generates multiple natural views, such as visual, textural, and semantic features. By simultaneously leveraging these multiple perspectives, multi-view clustering demonstrates superior robustness and accuracy compared to single-view approaches \cite{zhou2024survey}. 

The multi-view framework extends the conventional spectral clustering paradigm by introducing sophisticated matrix construction and graph fusion mechanisms while preserving the fundamental stages of graph learning, spectral embedding, and partition optimization. Information fusion in multi-view spectral clustering encompasses four principal strategies: (1) early fusion, which integrates view information prior to clustering \cite{hu2020multi, nie2016parameter, nie2017self, nie2017auto}, (2) late fusion, which combines individual view clustering outcomes \cite{liu2018late}, (3) hybrid fusion, which integrates information at multiple stages \cite{tan2020unsupervised}, and (4) score fusion, which aggregates view-specific clustering probabilities \cite{he2010performance}.

The choice of fusion method is critically dependent on the specific requirements of the clustering task, the nature of the multi-view data, and the desired balance between computational efficiency and clustering accuracy. Early fusion approaches, which create a cohesive representation of the data before applying clustering algorithms, often capture more intricate relationships between views and are the primary focus of many recent studies.
With the fusion strategy determined, the next crucial step in multi-view spectral clustering is the construction of appropriate matrices to represent the fused data.
\paragraph{Matrix Construction}
In multi-view spectral clustering, matrix construction offers two primary options - similarity matrix, directly constructed from the graph to represent pairwise similarities between data points across views, and Laplacian matrix, which encodes both similarity and degree information.
\paragraph{Graph Fusion}
Graph fusion, a critical step in multi-view spectral clustering, integrates information from multiple views into a unified representation. It comprises three main components:
\begin{itemize}
\item Consensus Information \cite{bai2024structural}: Identifies and emphasizes common patterns across different views, enhancing model robustness and generalization capability.
\item Complementary Information \cite{hao2023learning}: Leverages unique insights provided by each view, enriching the overall representation with diverse perspectives.
\item View Weighting \cite{zong2018weighted}: Assigns appropriate weights to each view, crucial for effective fusion. Strategies include equal-weighted, auto-weighted (parameter-weighted or parameter-free), and advanced approaches considering feature and sample weights, confidence levels, and cluster-wise weighting.
\end{itemize}
\begin{figure}[!h]
\centering
\includegraphics[width=1.0\linewidth]{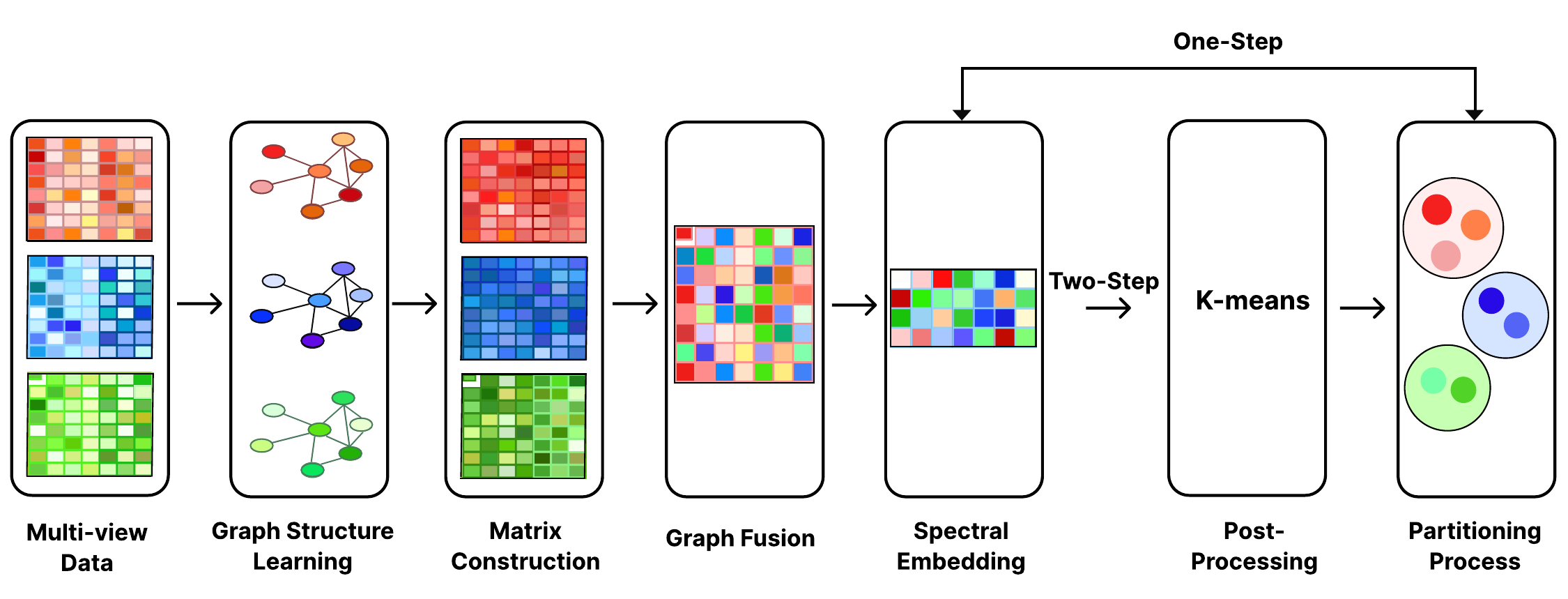}
\caption{Multi-view Spectral Clustering Framework.}
\label{Figure4}
\end{figure}
Figure \ref{Figure4} presents a systematic framework for multi-view spectral clustering, illustrating the sequential processes of graph construction, matrix formulation, information fusion, spectral embedding, and partition optimization. Building upon these foundational elements of multi-view spectral clustering, particularly the fusion mechanisms and matrix construction strategies, subsequent sections examine three principal graph learning paradigms: pairwise graphs, anchor-based representations, and hypergraph structures.

\subsubsection{Pairwise Graph}
Pairwise graph methods in multi-view spectral clustering focus on learning similarity graphs that capture cross-view relationships. Following the principles established in single-view clustering, these methods are categorized into fixed and adaptive approaches. While fixed methods employ predetermined rules, adaptive methods encompass three principal strategies: adaptive neighbor , initial similairy graph , and self-expressive learning. The subsequent sections analyze representative algorithms from each category, with self-expressive models examined separately due to their distinct theoretical foundations. Table  \ref{tab:anchor_graph_constructionmulti1} presents a systematic comparison of these methods, highlighting their graph learning mechanisms and fusion strategies.

\textbf{Fixed Methods:} Fixed methods use a predetermined number of neighbors for each data point. This category includes several representative methods. S-MVSC \cite{hu2020multi} uses KNN graphs and learns a sparse consensus similarity matrix directly without iterations. It doesn't model complementary information or view weights. The approach is two-step: first fusing graphs, then clustering with k-means. Its main advantage is speed and simplicity due to the closed-form solution. AMGL \cite{nie2016parameter} fuses Laplacian matrices from multiple views using automatically learned weights. It alternates between updating the fused Laplacian and the cluster indicator matrix, then applies k-means for final clustering. This parameter-free approach offers a simple yet effective solution for multi-view spectral clustering. SwMC \cite{nie2017self} learns a consensus graph from multiple views using Constrained Laplacian Rank. It automatically determines view weights without additional parameters. The method directly obtains clusters from the learned graph using Ky Fan theory, eliminating the need for post-processing like k-means. AMCSE \cite{shi2020auto} integrates multi-view graph learning, spectral embedding, and clustering into a unified framework. It adaptively weights view-specific Laplacian matrices, fuses them, and uses spectral rotation for direct discretization. This one-step approach jointly optimizes graph construction, view importance, and clustering, iteratively refining spectral embeddings and cluster assignments to improve multi-view clustering performance.

\textbf{Adaptive Neighbor Methods:} Adaptive neighbor methods dynamically adjust the neighborhood size based on data characteristics. MLAN \cite{nie2017auto} is a graph-based multi-view learning method that adaptively constructs a consensus similarity matrix across views. It automatically learns view weights without parameters, alternately optimizing the graph and weights. The final graph directly yields clusters via connected components, avoiding k-means and offering a parameter-free approach for clustering. GBS \cite{wang2019study} fuses multi-view similarity graphs into a consensus graph using auto-weighted fusion. It applies Ky Fan's theorem to constrain the Laplacian matrix, enabling direct one-step clustering without post-processing. This approach automatically weights views and produces clusters from the consensus graph.

GMC \cite{wang2019gmc} fuses multi-view similarity graphs into a unified graph, jointly optimizing view-specific and consensus graphs. It uses auto-weighted fusion and applies Ky Fan's theorem to constrain the Laplacian matrix, enabling direct one-step clustering. This approach mutually reinforces individual view graphs and the consensus graph, automatically weighting views without requiring post-processing. MVCGE \cite{el2022consensus} integrates multi-view consensus graph learning, spectral embedding, and nonnegative embedding (NESE) into a unified framework. It jointly optimizes these components using kernel matrices from multiple views, employing spectral convolution and automatic view weighting. This one-step approach aims to improve multi-view clustering by simultaneously refining graph structure, spectral representation, and cluster assignment without post-processing.

SCGLD \cite{zhong2022multi} integrates consensus graph learning and discrete clustering for multi-view data in a unified framework. It adaptively learns a shared graph across views while directly obtaining discrete cluster labels, avoiding information loss from separate steps. This one-step approach iteratively refines the consensus graph and cluster indicators, allowing them to mutually enhance each other, aiming to improve multi-view clustering performance. CSRF \cite{chen2023multiview} method fuses multi-view information at the spectral embedding level. It learns a consensus low-dimensional embedding by exploiting spectral rotation to combine individual view embeddings, reducing redundancy and capturing complementary information. CSRF uses an $\mathcal{O}(n^2)$ alternating optimization algorithm, incorporates flexible graph construction schemes, and constructs a fused similarity matrix from the consensus embedding for final clustering. This approach aims to improve multi-view clustering efficiency and effectiveness compared to methods operating on raw data or fixed graphs.

AGLSR \cite{tang2024one} is a one-step approach for multi-view clustering. It adaptively learns similarity graphs for each view, designs a shared spectral embedding across views, and uses spectral rotation to obtain clustering labels directly. AGLSR integrates graph learning, fusion, and partitioning into a unified framework, avoiding the need for separate k-means clustering on a consensus graph. This approach aims to improve clustering performance by closely unifying graph learning with partition generation.

\begin{table*}[!h]
\centering
\caption{Comparison of Fixed, Adaptive Neighbor, and Initial Similarity Matrix Multi-view Spectral Clustering Methods: Graph Construction and Clustering Steps}
\label{tab:anchor_graph_constructionmulti1}
  \scriptsize 
  \setlength{\tabcolsep}{1pt} 
  \renewcommand{\arraystretch}{1} 
\scalebox{0.85}{\begin{tabular}{|c|c|c|c|c|c|c|c|c|}
        \hline
        \multirow{3}{*}{\#} & \multirow{3}{*}{\textbf{Method}} & \multicolumn{1}{c|}{\multirow{3}{*}{\textbf{Graph Structure Learning}}} & \multicolumn{3}{c|}{\multirow{2}{*}{\textbf{Graph Fusion}}} & \multicolumn{3}{c|}{\textbf{Partitioning Process}} \\ \cline{7-9}
    &   &    \multicolumn{1}{c|}{}  &    \multicolumn{1}{c}{}     &    \multicolumn{1}{c}{}     &   & \multirow{2}{*}{\textbf{Two-Step}} & \multicolumn{2}{c|}{\textbf{One-Step}} \\ \cline{4-6}\cline{8-9}
   &   &  \multicolumn{1}{c|}{}     & \multicolumn{1}{c|}{\textbf{Consensus}} & \multicolumn{1}{c|}{\textbf{Complementary}} & \multicolumn{1}{c|}{\textbf{Weight}} & & \textbf{Continuous} & \textbf{Discretized} \\ \hline
1 & AMGL (2016) \cite{nie2016parameter} & KNN &  &  & \checkmark & \checkmark &  & \\
\hline
2 & S-MVSC (2018) \cite{hu2020multi} & KNN & \checkmark &  &  & \checkmark &  & \\
\hline
3 & SwMC (2017) \cite{nie2017self} & KNN & \checkmark &  &  &  & \checkmark & \\
\hline
4 & AMCSE (2020) \cite{shi2020auto} & KNN &  &  & \checkmark &  &  & \checkmark \\
\hline
5 & MLAN (2018) \cite{nie2017auto} & $\min\limits_{s_{ij}} \sum_{v=1}^V\alpha_v\sum_{i,j=1}^n (\Vert x_i^v - x_j^v\Vert_2^2 s_{ij} + \gamma s_{ij}^2)$ &  &  & \checkmark &  & \checkmark & \\
\hline
6 & MVGL (2018) & $\min\limits_{s_{ij}} \sum_{v=1}^V\alpha_v\sum_{i,j=1}^n (\Vert x_i^v - x_j^v\Vert_2^2 s_{ij} + \gamma s_{ij}^2)$ & \checkmark &  & \checkmark &  & \checkmark & \\
\hline
7 & OMSC (2019) & $\min\limits_{s_{ij}, t_{ij}, P, f} \sum_{i,j=1}^n \left( \|P^T x_i^v - P^T x_j^v\|_2^2 s_{ij} + \| f_i - f_j \|^2 t_{i,j} + \| s_{i}^{(v)} - t_{i} \|^2 \right)$ &  &  & \checkmark &  & \checkmark & \\

\hline
8 & GBS (2019) \cite{wang2019study} & $\min\limits_{s_{ij}} \sum_{v=1}^V\alpha_v\sum_{i,j=1}^n (\Vert x_i^v - x_j^v\Vert_2^2 s_{ij} + \gamma s_{ij}^2)$ & \checkmark &  & \checkmark &  & \checkmark & \\
\hline
9 & GMC (2020) \cite{wang2019gmc} & $\min\limits_{s_{ij}, P} \sum_{v=1}^V\alpha_v\sum_{i,j=1}^n (\Vert P^T x_i^v - P^T x_j^v\Vert_2^2 s_{ij} + \gamma s_{ij}^2)$ & \checkmark &  & \checkmark &  & \checkmark & \\
\hline
10 & MVCGE (2022) \cite{el2022consensus} & $\min\limits_{s_{ij}, P} \sum_{v=1}^V\alpha_v\sum_{i,j=1}^n (\Vert P^T x_i^v - P^T x_j^v\Vert_2^2 s_{ij} + \gamma s_{ij}^2)$ & \checkmark &  & \checkmark &  & \checkmark & \\
\hline
11 & SCGLD (2022) \cite{zhong2022multi} & $\min\limits_{s_{ij}} \sum_{v=1}^V\alpha_v\sum_{i,j=1}^n (\Vert x_i^v - x_j^v\Vert_2^2 s_{ij} + \gamma s_{ij}^2)$ & \checkmark &  & \checkmark &  &  & \checkmark \\
\hline
12 & CSRF (2023) \cite{chen2023multiview} & $\min\limits_{s_{ij}} \sum_{v=1}^V\alpha_v\sum_{i,j=1}^n (\Vert x_i^v - x_j^v\Vert_2^2 s_{ij} + \gamma s_{ij}^2)$ & \checkmark &  &  & \checkmark &  & \checkmark \\
\hline
13 & AGLSR (2023) \cite{tang2024one} & $\min\limits_{s_{ij}} \sum_{v=1}^V\alpha_v\sum_{i,j=1}^n (\Vert x_i^v - x_j^v\Vert_2^2 s_{ij} + \gamma s_{ij}^2)$ &  &  & \checkmark &  &  & \checkmark \\
\hline
\end{tabular}}
\end{table*}

\textbf{Self-expressive Methods}: Early developments in self-expressive multi-view spectral clustering focused on foundational representation learning and noise handling. As shown in Table \ref{tab:anchor_graph_constructionmulti2}, these methods can be analyzed through their graph structure learning, fusion strategies, and partitioning processes. DiMSC \cite{cao2015diversity} establishes a comprehensive framework integrating graph construction and fusion. It employs smooth representation with Frobenius norm $|X^v - X^vZ^v|_F^2$ and utilizes graph Laplacian regularization $\text{Tr}(Z^v L^v {Z^v}^T)$. Through the Hilbert-Schmidt Independence Criterion, it enhances view diversity and complementarity, providing efficient optimization with theoretical convergence guarantees via Sylvester equations. IMVSC \cite{wang2016iterative} advances this through a novel iterative framework combining multi-graph Laplacian regularization with low-rank and sparsity constraints. The method employs composite regularization terms $\|Z^v\|_* + \|Z^v\|_1 + \text{Tr}(Z^v L^v {Z^v}^T)$ with a robust $\ell_1$-norm reconstruction loss $|X^v - X^vZ^v|_1$. This approach effectively preserves view-specific structures while promoting cross-view consensus through alternating optimizationoptimization.

ECMSC \cite{wang2017exclusivity} introduces position-aware exclusivity terms with $\ell_1$-norm graph construction $\|Z^v\|_1$, enhancing complementary information while maintaining common structures. MLRSSC \cite{brbic2018multi} further advances subspace learning by simultaneously imposing nuclear and $\ell_1$-norm regularization $|Z^v| + \|Z^v\|_1$ with consensus constraints to learn a shared affinity matrix. CSMSC \cite{luo2018consistent} proposes a novel unified model exploiting both consistency and specificity through shared and view-specific representations $(C + D^v)$. Using $\ell_{2,1}$-norm reconstruction error $|X^v - X^v(C + D^v)|_{21}$ and composite regularization $|C|_* + |D^v|_F^2$, it effectively captures common properties and inherent differences across views. FMR \cite{li2019flexible} advances flexible representation learning using nuclear-norm regularization $|Z^v|$ with weighted HSIC measures, capturing nonlinear relationships while avoiding single-view limitations. GFSC \cite{kang2020multi} introduces adaptive graph fusion with Frobenius norm regularization $|X^v - X^vZ^v|_F^2 + \alpha|Z^v|_F^2$, maintaining optimal clustering properties through explicit structure constraints.

Recent developments have focused on sophisticated integration of local-global structures and advanced fusion strategies. MSCCSM \cite{hao2021multi} maximizes the common structure between local and global representations through a dynamic learning framework. The method employs $\ell_2$-norm regularization $|W|_F^2$ combined with global reconstruction loss $|X^v - X^vZ|_F^2$, while utilizing a structure fusion term $|Z \odot W|_1$ to enhance view complementarity. This unified approach enables better preservation of both local manifold structure and global clustering information. DMSC-UDL \cite{wang2020deep} represents a significant advancement by leveraging deep neural networks for graph construction. The method uniquely combines manifold structure preservation through $\text{Tr}({H^v}^T L H^v)$ with discriminative cross-view constraints $|S_i \otimes S_j|_1$. This deep learning framework effectively captures nonlinear relationships while maintaining complementary information across views, leading to more discriminative clustering results.

COMVSC \cite{zhang2020consensus} introduces a novel one-step framework that unifies affinity learning, partition fusion, and clustering. Through subspace representation construction $|X^v - X^vZ^v|_F^2 + \alpha|Z^v|_F^2$ and partition-level fusion, it avoids the limitations of noisy affinity matrices. Notably, it employs spectral rotation for direct cluster assignment, eliminating the need for additional post-processing steps and ensuring more coherent clustering results. MVSCGE \cite{chen2020multiview} advances locality preservation in learned subspaces through a comprehensive approach combining grouping effects with cross-view consistency. The method's smooth regularization $\lambda\text{tr}(Z^{(v)}L^{(v)}{Z^{(v)}}^T)$ ensures robust local structure preservation, while its unified optimization framework jointly learns subspace representations and consistent cluster indicators across views, effectively balancing local and global information. MAGC \cite{lin2021multi} presents an innovative framework integrating multiple graph structures and node attributes through sophisticated graph filtering. Unlike deep learning approaches, it employs efficient graph filtering $|X^v - X^vS|_F^2$ combined with weighted high-order random walk proximity $|Z - f(A^v)|_F^2$. This design enables smooth node representations while adaptively learning consensus relationships across heterogeneous views. The method's flexible regularizer captures complex relationships across views, making it particularly effective for handling multi-view attributed graphs.

\begin{table*}[!h]
\centering
\caption{Comparison of Self-expressive Multi-view Spectral Clustering Methods: Graph Construction and Clustering Steps}
\label{tab:anchor_graph_constructionmulti2}
\scriptsize
\setlength{\tabcolsep}{1pt}
\renewcommand{\arraystretch}{1}
\begin{tabular}{|c|c|c|c|c|c|c|c|c|c|}
\hline
\multirow{3}{*}{\#} & \multirow{3}{*}{\textbf{Method}} & \multicolumn{2}{c|}{\multirow{2}{*}{\textbf{Graph Structure Learning}}} & \multicolumn{3}{c|}{\multirow{2}{*}{\textbf{Graph Fusion}}} & \multicolumn{3}{c|}{\textbf{Partitioning Process}} \\ \cline{8-10}
& & \multicolumn{1}{c}{} & \multicolumn{1}{c|}{} & \multicolumn{1}{c}{} & \multicolumn{1}{c}{} & \multicolumn{1}{c|}{} & \multirow{2}{*}{\textbf{Two-Step}} & \multicolumn{2}{c|}{\textbf{One-Step}} \\ \cline{3-7}\cline{9-10}
& & \multicolumn{1}{c|}{$\textbf{Loss}(X,Z)$} & \multicolumn{1}{c|}{\textbf{Regularization}} & \multicolumn{1}{c|}{\textbf{Consensus}} & \multicolumn{1}{c|}{\textbf{Complementary}} & \multicolumn{1}{c|}{\textbf{Weight}} & & \textbf{Continuous} & \textbf{Discretized} \\ \hline

 1 & DiMSC (2015) \cite{cao2015diversity} & $\|X^v - X^vZ^v\|_F^2$ & $Tr(Z^v L^v {Z^v}^T)$ & & \checkmark & \checkmark & \checkmark & & \\
\hline
2 & IMVSC (2016) \cite{wang2016iterative} & $\|X^v - X^vZ^v\|_1$ & $\|Z^v\|_* + \|Z^v\|_1 + Tr(Z^v L^v {Z^v}^T)$ & \checkmark & & & \checkmark & & \\
\hline
3 & ECMSC (2017) \cite{wang2017exclusivity} & $\|X^v - X^vZ^v\|_1$ & $\|Z^v\|_1$ & \checkmark & \checkmark & & \checkmark & & \\
\hline

 4 & MLRSSC (2018) \cite{brbic2018multi} & $\|X^v - X^vZ^v\|_F^2$ & $\|Z^v\|_* + \|Z^v\|_1$ & \checkmark & & & \checkmark & & \\
\hline
5 & CSMSC (2018) \cite{luo2018consistent} & $\|X^v - X^v(C + D^v)\|_{21}$ & $\|C\|_* + \|D^v\|_F^2$ & \checkmark & \checkmark & & \checkmark & & \\
\hline
6 & FMR (2019) \cite{li2019flexible} & $\|X^v - X^vZ^v\|_{21}$ & $\|Z^v\|_*$ & & \checkmark & & \checkmark & & \\
\hline
7 & GFSC (2020) \cite{ kang2020multi} & $\|X^v - X^vZ^v\|_F^2$ & $\|Z^v\|_F^2$ & \checkmark & & & \checkmark & & \\
\hline
8 & MSCCSM (2021) \cite{hao2021multi} & $\|X^v - X^vZ^v\|_F^2$ & $\|Z^v\|_F^2$ & \checkmark & & & \checkmark & & \\
\hline
9 & DMSC-UDL (2021) \cite{wang2020deep} & $\|X^v - \hat{X^v}\|_F^2 + \|H^v - H^vZ\|_F^2$ & $Tr({H^v}^T L H^v) + \|Z^v\|_F^2$ & & \checkmark & & & \checkmark & \\
\hline
10 & COMVSC (2022) \cite{zhang2020consensus} & $\|X^v - X^vZ^v\|_F^2$ & $\|Z^v\|_F^2$ & \checkmark & & \checkmark & & & \checkmark \\
\hline
11 & MvSCGE (2022) \cite{chen2020multiview} & $\|X^v - X^vZ^v\|_F^2$ & $Tr(Z^v L^v {Z^v}^T)$ & \checkmark & & & \checkmark & & \\
\hline
12 & MAGC (2023) \cite{lin2021multi} & $\|X^v - X^vZ\|_F^2$ & $\|Z - f(A^v)\|_F^2$ & & & \checkmark & \checkmark & & \\
\hline

\end{tabular}
\end{table*}

\subsubsection{Anchor Graph}  
Anchor graph-based methods in multi-view spectral clustering offer a significant advancement in managing scalability challenges. By using a small set of representative points (anchors), these methods approximate the full similarity structure, reducing computational complexity while preserving critical data relationships. MVSC \cite{li2015large} selects anchors through k-means on concatenated features. It employs a fixed method to generate anchor graphs and constructs bipartite graphs for similarity matrices. The method uses a weighted combination for view fusion and performs spectral embedding on the fused graph to obtain low-dimensional representations, followed by k-means clustering to derive the final data partitions. DSC \cite{luo2019discrete} adopts k-means for anchor selection and constructs adaptive anchor graphs. It generates similarity matrices via bipartite graphs, incorporates view-specific weights automatically, and achieves clustering through a single-step, continuous optimization process that directly learns a consistent partition across views.

FMDC \cite{qiang2021fast} utilizes k-means++ for anchor point selection and adopts an adaptive method for anchor graph generation. The similarity matrices are computed using the Gram matrix approach. Views are combined using learned weights, and clustering is achieved through a single-step, discretized optimization process that directly learns the cluster indicator matrix across views. SFMC \cite{li2020multiview} selects anchor points using k-means++ and employs an adaptive method for anchor graph construction. It generates similarity matrices through bipartite graphs and achieves clustering in a single-step, continuous optimization process, directly learning a consistent partition across views using Ky Fan's theorem. Similarly, BIGMC \cite{li2020bipartite} selects anchors via k-means and constructs adaptive anchor graphs using bipartite graphs between data points and anchors. It incorporates consensus information, learns view-specific weights automatically, and performs clustering through a single-step, continuous optimization process, directly learning a joint graph compatible across views.

ERMC-AGR \cite{yang2022efficient} and ECMC \cite{yang2022efficient} both use k-means for anchor selection and fixed methods for anchor graph construction. ERMC-AGR computes similarity matrices using the Gram matrix approach, incorporates consensus information across views, and merges views with learned weights. The clustering process involves two stages: spectral embedding on the fused graph followed by k-means clustering for final partitioning. On the other hand, ECMC generates similarity matrices using bipartite graphs and automatically assigns weights during view fusion. It achieves clustering through a single-step, discretized optimization process that directly learns the cluster indicator matrix. FMCSE \cite{yang2022fast} also utilizes k-means for anchor point selection but adopts an adaptive method for anchor graph generation. Similarity matrices are computed using the Gram matrix approach, and views are combined with learned weights. The method performs clustering through a single-step, discretized optimization process, directly learning the cluster indicator matrix.

E$^2$OMVC \cite{wang2023efficient} and FMAGC \cite{yang2024fast} follow similar methodologies with slight variations. \(E^2OMVC\) employs k-means for anchor selection, uses a fixed method for anchor graph construction, and generates similarity matrices using the Gram matrix approach. It incorporates consensus information across views, merges views using learned weights, and performs clustering through a single-step, discretized optimization process that directly learns the cluster indicator matrix. FMAGC uses k-means++ for anchor selection, employs an adaptive method for anchor graph construction, and computes similarity matrices using the Gram matrix approach. It incorporates consensus information across views and combines them through a weighted approach. The clustering process involves a single-step, discretized optimization that directly learns the cluster indicator matrix. Table \ref{tab:anchor_graph_constructionmulti3} provides a comparative overview of these methods, summarizing their approaches to graph construction and clustering steps.

\begin{table*}[!h]
  \centering
  \caption{Comparison of Anchor Graph Multi-view Spectral Clustering Methods: Graph Construction and Clustering Steps}
  \label{tab:anchor_graph_constructionmulti3}
    \scriptsize 
  \setlength{\tabcolsep}{1pt} 
  \renewcommand{\arraystretch}{1} 
\begin{tabular}{|c|c|c|c|c|c|c|c|c|c|c|c|}
        \hline
        \multirow{3}{*}{\#} & \multirow{3}{*}{\textbf{Method}} & \multicolumn{3}{c|}{\multirow{2}{*}{\textbf{Graph Structure Learning}}} & \multicolumn{3}{c|}{\multirow{2}{*}{\textbf{Graph Fusion}}} & \multicolumn{3}{c|}{\textbf{Partitioning Process}} \\ \cline{9-11}
    &  &   \multicolumn{1}{c}{}     &  \multicolumn{1}{c}{}     & \multicolumn{1}{c|}{}     &    \multicolumn{1}{c}{}  &    \multicolumn{1}{c}{}     &    \multicolumn{1}{c|}{}     &  \multirow{2}{*}{\textbf{Two-Step}} & \multicolumn{2}{c|}{\textbf{One-Step}} \\ \cline{3-8}\cline{10-11}
   &   &  \multicolumn{1}{c|}{\textbf{Select Anchor}} & \multicolumn{1}{c|}{\textbf{Anchor Graph}}  & \multicolumn{1}{c|}{\textbf{Similarity Matrix}}   & \multicolumn{1}{c|}{\textbf{Consensus}} & \multicolumn{1}{c|}{\textbf{Complementary}} & \multicolumn{1}{c|}{\textbf{Weight}} & & \textbf{Continuous} & \textbf{Discretized} \\ \hline
    1 & MVSC (2015) \cite{li2015large} & K-means & Fixed & Bipartite & & & \checkmark & & & \\
    \hline
    2 & DSC (2019) \cite{luo2019discrete} & K-means & Fixed & Gram & &  & \checkmark & & & \checkmark \\
    \hline
    3 & FMDC (2021) \cite{qiang2021fast} & K-means++ & Adaptive & Gram & & & \checkmark &  & & \checkmark \\
    \hline
    4 & SFMC (2022) \cite{li2020multiview} & K-means & Fixed & Bipartite &  & & \checkmark & & \checkmark & \\
    \hline
    5 & BIGMC (2022) \cite{li2020bipartite} & K-means & Adaptive & Bipartite & \checkmark & & \checkmark & & \checkmark & \\
    \hline
    6 & ERMC-AGR (2022) \cite{yang2022efficient} & K-means & Fixed & Gram & \checkmark & & \checkmark & \checkmark &  & \checkmark  \\
    \hline
    7 & ECMC (2022) \cite{yang2022efficient} & K-means & Fixed & Bipartite & \checkmark  &  & \checkmark & &  & \checkmark \\
    \hline
    8 & FMCSE (2022) \cite{yang2022fast} & K-means & Adaptive & Gram & & & \checkmark & & & \checkmark \\
    \hline
    9 & E$^2$OMVC (2023) \cite{wang2023efficient} & K-means & Fixed & Gram & \checkmark & & \checkmark & & & \checkmark \\
    \hline
    10 & FMAGC (2024) \cite{yang2024fast} & K-means++ & Adaptive &  Gram &  & & \checkmark & & & \checkmark \\
    \hline
  \end{tabular}
\end{table*}

\subsubsection{Hypergraph}
Hypergraph-based methods for multi-view spectral clustering remain underexplored, with limited research addressing this promising direction. Hao et al. \cite{hao2020self} introduced a hypergraph-based method that adaptively weighs views while capturing higher-order relationships. The approach constructs hypergraph Laplacians to encode high-order relations and employs a self-weighting strategy to determine view importance. By integrating hypergraph regularization with self-weighting in a self-representation framework, it learns an optimal consensus representation, preserving both view-specific and high-order information, followed by spectral clustering for partitioning. Yang et al. \cite{yang2023hypergraph} proposed an adaptive hypergraph learning method to model higher-order relationships in multi-view data. This method learns view-specific similarity matrices and balances shared and view-specific information using a weighted fusion strategy, leading to improved clustering via a two-step process. Chen et al. \cite{chen2024incomplete} developed IMVC-HG for incomplete multi-view clustering. The method generates view-specific affinity matrices through self-representation, constructs a consistent hypergraph, and integrates representation learning and clustering using non-negative matrix factorization with orthogonal constraints in a unified framework, capturing both consensus and view-specific information.

\section{Future Research Direction} \label{future}

The field of single-view and multi-view spectral clustering has seen remarkable advancements in recent years, yet it still offers a promising outlook for future research. As data continues to grow in complexity, diversity, and scale, there is a critical need to develop methods that are not only robust and adaptive but also efficient in handling such challenges. Below, we outline key directions for future research, emphasizing enhancements that could significantly improve the effectiveness and applicability of single-view and multi-view spectral clustering across diverse domains:

\subsection{Finding Good Neighbors for Graph Construction}  
Future research should prioritize the development of advanced Good Neighbor strategies that identify the most representative and relevant neighbors for each data point. Ensuring the similarity graph accurately captures the underlying data structure is essential for effective clustering. This could involve investigating new criteria or metrics for neighbor selection that go beyond traditional distance-based methods. Additionally, incorporating domain-specific knowledge or machine learning techniques can further refine this selection process, enabling the construction of more reliable and context-aware similarity graphs.

\subsection{Graph Construction Using Similarity and Dissimilarity Information}  
Incorporating both similarity and dissimilarity information into adaptive neighbor selection processes offers a promising direction for enhancing clustering performance. By dynamically adjusting neighbors based on both positive and negative relationships between data points, future methods could achieve greater robustness and accuracy. This approach may benefit from advancements in metric learning or semi-supervised learning, allowing for the guided construction of more meaningful and discriminative graphs. Exploring the interplay between similarity and dissimilarity in graph construction could yield more nuanced and flexible clustering outcomes.

\subsection{High-Order Adaptive Neighbor Selection}  
High-order relationships, such as those represented by motifs or hypergraphs, provide richer insights into the structural characteristics of data compared to simple pairwise connections. Future research could focus on developing high-order adaptive neighbor selection methods that consider not only direct connections but also the broader relational context in which these connections occur. This may involve constructing higher-dimensional similarity matrices to capture intricate interactions among data points, resulting in more precise and interpretable clustering results. Such advancements could significantly enhance the ability of clustering methods to capture the true complexity of data structures.

\subsection{Semi-Supervised Multi-View Spectral Clustering}  
The integration of semi-supervised techniques into spectral clustering holds significant potential for improving performance by leveraging limited labeled data. Future efforts could aim to develop semi-supervised multi-view spectral clustering methods that effectively combine labeled and unlabeled data across multiple views. This could include designing innovative regularization terms or loss functions that incorporate label information, as well as creating novel strategies to propagate labels through the spectral embedding process. Such methods could bridge the gap between supervised and unsupervised approaches, leading to more informed and accurate clustering results.

\subsection{Multi-View Spectral Clustering Using Adaptive Hypergraphs}  
Hypergraphs represent an advanced framework for capturing complex multi-way relationships within data. Future research could explore adaptive hypergraph-based methods for multi-view spectral clustering, where the hypergraph structure dynamically adjusts to the characteristics of the data. This may involve developing novel techniques for constructing hypergraphs that account for the varying importance of different views, or creating efficient algorithms to process these structures for clustering purposes. By harnessing the adaptability and expressiveness of hypergraphs, such methods could deliver more accurate and insightful clustering outcomes, especially in scenarios involving multi-view data.

\section{Conclusion} \label{conclusion}

This paper has presented a systematic survey of spectral clustering methods, focusing on the fundamental role of Graph Structure Learning (GSL) in improving clustering performance. We have examined diverse graph construction approaches, from pairwise and anchor-based methods to hypergraph techniques, analyzing both fixed and adaptive implementations. Our review encompasses single-view and multi-view frameworks, including their applications in one-step and two-step clustering processes, along with effective multi-view information fusion strategies. Through extensive analysis of various graph construction techniques and their impact on clustering accuracy, we highlight the importance of GSL in handling high-dimensional and large-scale datasets. By providing this comprehensive analysis of current methods and emerging challenges, we aim to guide future research in spectral clustering, particularly in developing efficient graph construction techniques and scalable solutions for large-scale data applications. This work contributes to the advancement of spectral clustering as an essential tool in modern data analysis and machine learning.

 \ifCLASSOPTIONcaptionsoff
\newpage
\fi

\bibliographystyle{IEEEtran}
\bibliography{main}

\vspace{-1.4 cm}
\begin{IEEEbiography}[{\includegraphics[width=1in,height=1in,clip,keepaspectratio]{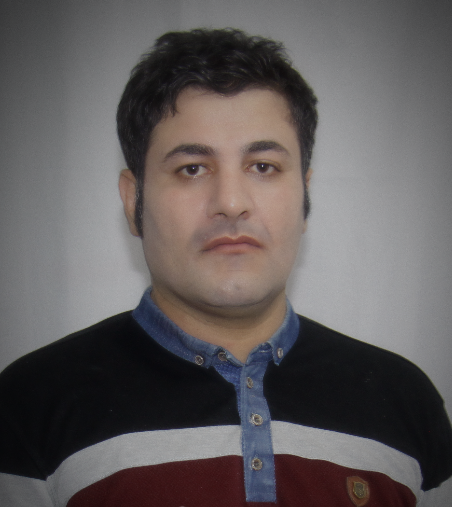}}]{Kamal Berahmand} 
received his Ph.D. in 2024 from the School of Computer Science, Queensland University of Technology (QUT), Australia, where he is now a researcher. His current research interests include machine learning, representation learning, graph neural networks, and graph learning.
\end{IEEEbiography}

\vspace{-1.45cm}
\begin{IEEEbiography}[{\includegraphics[width=1in,height=1.25in,clip,keepaspectratio]{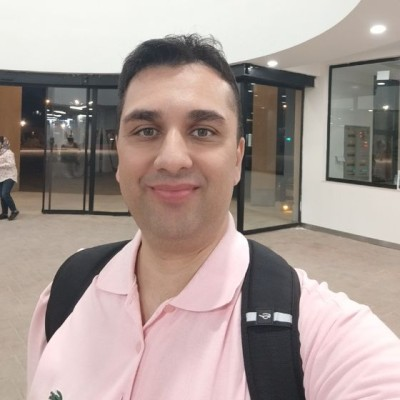}}]{Farid Saberi-Movahed} earned his Ph.D. in Applied Mathematics, specializing in Optimization and Numerical Linear Algebra, from Shahid Bahonar University of Kerman, Iran. Currently, he serves as an Associate Professor in the Department of Applied Mathematics, Faculty of Sciences and Modern Technologies, at the Graduate University of Advanced Technology, Kerman, Iran. His research interests focus on numerical linear algebra methods, with an emphasis on their applications in Data Mining and Machine Learning.
\end{IEEEbiography}

\vspace{-1.4cm}
 \begin{IEEEbiography}[{\includegraphics[width=1in,height=1.25in,clip,keepaspectratio]{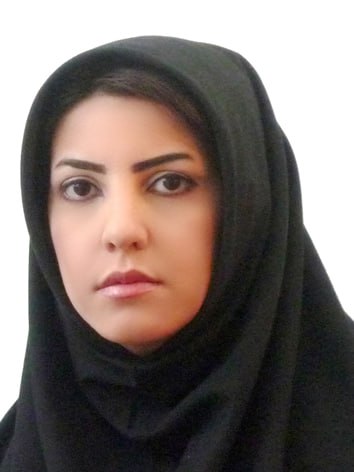}}]{Razieh Sheikhpour} received her Ph.D. in Computer
Engineering from Yazd University, Yazd, Iran, in
2017. Currently, she is an Associate Professor at the
department of Computer Engineering at Ardakan
University, Ardakan, Iran. Her research interests
include machine learning, semi-supervised feature
selection and bioinformatics.
\end{IEEEbiography}

\vspace{-1.4cm}
\begin{IEEEbiography}[{\includegraphics[width=1in,height=1.25in,clip,keepaspectratio]{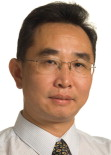}}]
{Yuefeng Li}  is currently a professor and the HDR Director with the School of Computer Science, Queensland University of Technology (QUT), Australia. His work has received 7,146 citations, with an overall h-index of 41 and an i10 index of 138. He has published 12 articles with more than 100 citations and 5 articles with more than 200 citations. Professor Yuefeng Li carries out research on data mining, text analysis, and AI-based Data Analysis. He has a lot of experience in developing new data mining and machine learning algorithms in all data types (structured and unstructured). He is recognized internationally in Knowledge \& Data Engineering and Natural Language Processing. He has demonstrable experience in leading large-scale research projects and has achieved many established research outcomes that have been published and highly cited in top data mining journals and conferences, such as the TKDE, ACM Trans. Knowl. Discov. Data, Data Min. Knowl. Discov., IEEE Trans. Netw. Sci. Eng., Information Sciences, Artificial Intelligence Review, KDD, CIKM, WWW, ICDM, and PAKDD. He is an Editor-in-Chief of Web Intelligence journal.

\end{IEEEbiography}
\vspace{-1.4cm}
\begin{IEEEbiography}[{\includegraphics[width=1in,height=1.25in,clip,keepaspectratio]{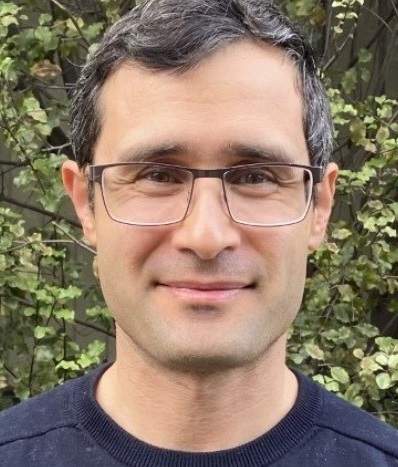}}]{Mahdi Jalili} (Senior Member, IEEE) received the B.S. degree in electrical engineering from Tehran Polytechnic, Tehran, Iran, in 2001, the M.S. degree in electrical engineering from the University of Tehran, Tehran, in 2004, and the Ph.D. degree in synchronization in dynamical networks from the Swiss Federal Institute of Technology Lausanne, Lausanne, Switzerland. He then joined the Sharif University of Technology as an Assistant Professor. He is currently a Senior Lecturer with the School of Engineering, RMIT University, Melbourne, VIC, Australia. His research interests include network science, dynamical systems, social networks analysis and mining, and data analytics. Dr. Jalili was a recipient of the Australian Research Council DECRA Fellowship and the RMIT Vice-Chancellor Research Fellowship. He is an Associate Editor of the IEEE Canadian Journal of Electrical and Computer Engineering and an Editorial Board Member of the Mathematical Problems in Engineering and Complex Adaptive Systems Modeling.
\end{IEEEbiography}

\end{document}